\documentclass{article}

\PassOptionsToPackage{numbers, compress}{natbib}

\usepackage[main, final]{neurips_2026}

\usepackage[utf8]{inputenc} %
\usepackage[T1]{fontenc}    %
\usepackage{hyperref}       %
\usepackage{url}            %
\usepackage{booktabs}       %
\usepackage{amsfonts}       %
\usepackage{nicefrac}       %
\usepackage{microtype}      %
\usepackage[table]{xcolor}         %

\usepackage{graphicx}
\usepackage{tikz}
\usepackage{caption}
\usepackage{subcaption}
\usepackage{multirow}
\usepackage{wrapfig}
\usepackage[most]{tcolorbox}

\usepackage{amsmath}
\usepackage{amssymb}
\usepackage{mathtools}
\usepackage{amsthm}

\usepackage{algorithm,algpseudocode}
\usepackage[capitalize]{cleveref}

\theoremstyle{plain}

\theoremstyle{definition}

\theoremstyle{remark}

\usepackage{xcolor}
\usepackage{xspace}
\newcommand{\eg}{\textit{e.g.}\xspace}
\newcommand{\ie}{\textit{i.e.}\xspace}

\usepackage{pifont}         %
\definecolor{darkergreen}{RGB}{21, 152, 56}
\definecolor{red2}{RGB}{252, 54, 65}
\newcommand{\yesmark}{\textcolor{darkergreen}{\ding{52}}}
\newcommand{\nomark}{\textcolor{red2}{\ding{56}}}

\newcommand{\renyient}{R\'enyi entropy\xspace}
\newcommand{\renyitwoent}{R\'enyi-2 entropy\xspace}
\newcommand{\renyitwo}{R\'enyi-2\xspace}

\newcommand{\methodname}{\textsc{CoMet}\xspace}
\newcommand{\methodnamefull}{Context and Multiplicity Decomposition for Uncertainty Estimation\xspace}
\newcommand{\methodnamefullhighlight}{\textbf{Co}ntext and \textbf{M}ultiplicity D\textbf{e}composi\textbf{t}ion (\ours) for Uncertainty estimation\xspace}
\newcommand{\ours}{\methodname}
\newcommand{\oursfull}{\methodnamefull}

\crefname{theorem}{Thm.}{Thms.}
\Crefname{theorem}{Theorem}{Theorems}

\crefname{lemma}{Lem.}{Lems.}
\Crefname{lemma}{Lemma}{Lemmas}

\crefname{definition}{Def.}{Defs.}
\Crefname{definition}{Definition}{Definitions}

\crefname{assumption}{Asm.}{Asms.}
\Crefname{assumption}{Assumption}{Assumptions}

\crefname{section}{Sec.}{Secs.}
\Crefname{section}{Section}{Sections}

\crefname{figure}{Fig.}{Figs.}
\Crefname{figure}{Figure}{Figures}

\crefname{table}{Tab.}{Tabs.}
\Crefname{table}{Table}{Tables}

\crefname{appendix}{Appendix}{Appendices}
\Crefname{appendix}{Appendix}{Appendices}

\definecolor{codegreen}{rgb}{0,0.6,0}
\definecolor{codegray}{rgb}{0.5,0.5,0.5}
\definecolor{codepurple}{rgb}{0.58,0,0.82}
\definecolor{backcolour}{rgb}{0.95,0.95,0.92}

\lstdefinestyle{mystyle}{
  backgroundcolor=\color{backcolour}, commentstyle=\color{codegreen},
  keywordstyle=\color{magenta},
  numberstyle=\tiny\color{codegray},
  stringstyle=\color{codegreen},
  basicstyle=\ttfamily\footnotesize,
  breakatwhitespace=false,         
  breaklines=true,                 
  captionpos=b,                    
  keepspaces=true,                 
  numbers=left,                    
  numbersep=5pt,                  
  showspaces=false,                
  showstringspaces=false,
  showtabs=false,                  
  tabsize=2
}

\title{\ours: Context and Multiplicity Decomposition for Multimodal Uncertainty Estimation}

\author{
Sanghyuk Chun \quad William Yang \quad Amaya Dharmasiri \quad Olga Russakovsky\\
\\
Princeton University
}

\begin{document}

\maketitle

\begin{abstract}
Uncertainty estimation has been a long-standing challenge in AI models; it amounts to ``knowing what you don't know,'' and metacognition is notoriously difficult even for humans (cf. the Dunning-Kruger effect). Although it is still far from solved even in simpler classification systems, tackling it in multimodal large language models (MLLMs) is becoming increasingly important. Within MLLMs, uncertainty can stem from any of the diverse sources as well as from their relationships, and further can stem from the unbounded answers in the open-ended setting.
To tackle the issues, we propose \ours, an MLLM uncertainty estimation method by decomposing uncertainty into a context-specific term and a multiplicity-specific term. The former captures ambiguity induced by the given context (\eg, task or prompt), while the latter captures how many plausible answers determined by the context remain compatible with the given input. We train a lightweight post-hoc uncertainty module to estimate these quantities, which enables efficient uncertainty estimation without autoregressive answer generation or repeated sampling. Experiments on various open-ended multimodal benchmarks, hallucination detection, and multiple-choice visual question answering benchmarks show that \ours consistently improves uncertainty estimation over existing baselines while remaining efficient in practice.
Code is available at \url{https://github.com/princetonvisualai/comet_uncertainty}
\end{abstract}

\section{Introduction}

Multimodal large language models (MLLMs) are increasingly used in complex and open-ended settings where inputs are diverse and outputs are not always verifiable. Consequently, reliable uncertainty estimation becomes critical for downstream decision-making: beyond solely producing an answer, a model should also assign a score that reflects the ambiguity of the prediction for a given input. This score can be used for self-evaluation (whether a prediction is correct) \cite{kadavath2022p_correct,lin2022teaching,verbalized_confidence,xiong2024verbalized_confidence_iclr,xiao2026vl_calibration}, abstention (whether a model can correctly answer the given input) \cite{rajpurkar2018know,whitehead2022reliable,zhang2024r,wen2025know}, self-improvement \cite{wang2022self_consistency}, or hallucination detection \cite{zhang2024vl_uncertainty,farquhar2024semantic_entropy,park2026glsim}. However, uncertainty estimation in MLLMs is challenging due to two fundamental properties: the diversity of uncertainty sources and the open-ended nature of the answer space. 

First, uncertainty can arise from each modality as well as from their many-to-many relationship. Let us take a vision-language task as an example. Visual uncertainty can arise from low image quality \cite{peterson2019human,shi2019probabilistic}, \eg, a blurry image is inherently more uncertain for a model. The text modality can also be intrinsically ambiguous when the text itself maps to multiple semantic meanings \cite{vilnis2014word}, \eg, ``What color is the mouse?'' may refer either to the animal or to the computer equipment. Moreover, even when both the input image and the input text are individually certain, the many-to-many nature of the input-output correspondence (\ie, their multiplicity \cite{chun2025multiplicity}) can still introduce ambiguity \cite{chun2021pcme,chun2024pcmepp,chun2025prolip}. For example, an image with multiple objects and a question ``What is in this image?'' can be mapped to multiple plausible textual descriptions \cite{chun2022eccv_caption}. As a result, uncertainty in multimodal tasks is inherently compositional, and a faithful uncertainty estimate should distinguish different sources of ambiguity.

\begin{figure}[t]
    \centering
    \includegraphics[width=\linewidth]{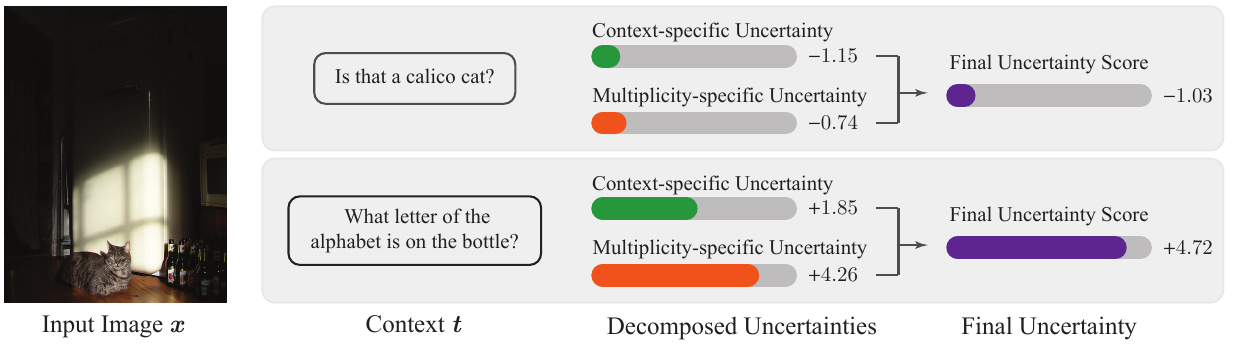}
    \caption{\small {\bf Multimodal uncertainty decomposition.} Even for the same image $x$, uncertainty could vary by the given context $t$. We decompose the uncertainty of a multimodal input into two components: (1) \textbf{Context-specific uncertainty} quantifies how broadly the context $t$ defines the answer candidates (\eg, ``is that a calico cat?'' has two valid answer candidates, ``yes'' and ``no'', whereas ``what letter of the alphabet is on the bottle?'' has 26 candidates)
    (2) \textbf{Multiplicity-specific uncertainty} quantifies how many answer candidates (\eg, ``yes'' and ``no'') remain plausible with the input $x$ (\eg, the cat certainly corresponds to a tabby cat, and therefore ``no'' is the only plausible answer, whereas the bottles are small and not well-lit, thus increasing rather than reducing uncertainty)
    }
    \label{fig:uncertainty_teaser}
\vspace{-1em}
\end{figure}

Second, the set of valid answers is unbounded in open-ended settings. Unlike closed-set tasks, where uncertainty can be measured over a fixed candidate set using quantities such as entropy \cite{kendall2017uncertainties} or logit confidence \cite{hendrycks2016baseline}, open-ended tasks do not provide an explicit answer space to compare against. 
Instead, recent studies on LLM uncertainty quantification often utilize the generative capability of LLMs, for example, by asking a model to assess its confidence \cite{kadavath2022p_correct,verbalized_confidence} or approximating the open-ended answer space with a finite set of generated answers \cite{farquhar2024semantic_entropy,wang2022self_consistency}. However, these approaches rely heavily on generation performance, are sensitive to generation hyperparameters, and introduce an additional inference cost. In practice, these methods often require additional uncertainty-aware instruction tuning \cite{kapoor2024large,damani2026beyond} on the target LLM, which further increases training and deployment cost.

We aim to estimate the predictive uncertainty for MLLMs by considering both (1) the compositional nature of multimodal uncertainty and (2) the unbounded answer space of open-ended settings, without relying on the LLM generation capability at inference time or expensive model fine-tuning. To address these challenges, we propose a novel decomposition of predictive uncertainty into a context-specific term and a multiplicity-specific term. The first term captures the ambiguity induced by context $t$, where $t$ determines the set of answer candidates. For example, in visual question answering (VQA) tasks, $t$ is a text question.
\cref{fig:uncertainty_teaser} shows an example: the question ``What letter of the alphabet is on the bottle?'' (26 candidates) induces a higher uncertainty than ``Is that a calico cat?'' (2 candidates). The second term captures how many candidates remain plausible with the given input $x$. For example, the cat in the image is a tabby cat; therefore, only ``No'' remains plausible when $x$ is given.

We construct our uncertainty measure using an answer candidate $y$ for the given context $t$ and a new binary variable $m$. A candidate answer $y$ is matched ($m=1$) if it is semantically valid when question $t$ is asked about image $x$. 
For example, let $x$ be an image of a cat and $t$ be the question ``What is in this image?''. The answer candidate $y=$``a cat'' is matched ($m=1$), while $y=$``a dog'' is not ($m=0$).
Using this variable, we define a match-conditioned posterior $p(y \mid x, t, m=1)$ and quantify its uncertainty using entropy. However, in open-ended settings, standard discrete entropy is not directly applicable because the answer space is not explicitly enumerable and can be viewed as continuous. To tackle the issue, we employ \renyitwoent, a generalized entropy measure.
Our derivation represents uncertainty 
in terms of matching probability $p(m=1\mid x,t,y)$ (\ie, ``is $y$ a plausible answer of $x, t$?'') without autoregressive answer generation or repeated sampling from $p_\theta(y\mid x,t)$. This avoids sensitivity to various generation hyperparameters and design choices.

In practice, we estimate the matching probability $p(m=1 \mid x,t,y)$ using an MLLM-as-verifier strategy and approximate $p(y \mid t)$ with answers collected from visual instruction tuning datasets. This provides a finite candidate set without autoregressive candidate generation. Using these two ingredients, we train lightweight post-hoc uncertainty modules that predict our uncertainty formulation from an input $x$ and a context $t$. Once trained, our uncertainty estimation module requires only $x$ and $t$ at inference time, without access to the candidate-answer set or any generation process.

We evaluate our uncertainty estimation framework, named \methodnamefullhighlight, 
on diverse open-ended multimodal benchmarks, including VQA v2 \cite{vqav2}, VizWiz \cite{vizwiz}, and OK-VQA \cite{okvqa}. We additionally evaluate our method on a hallucination detection benchmark \cite{hallusionbench} and multiple choice VQA benchmarks, such as MMMU \cite{mmmu}, MMMU Pro \cite{mmmu_pro}, and MMStar \cite{mmstar}. We validate \ours across multiple backbones, including Qwen3VL 2B, 4B, 8B Instruct \cite{bai2025qwen3}, and InternVL3.5-1B \cite{wang2025internvl3_5}. 
Across these settings, \ours more reliably identifies inputs that are likely to be incorrect than existing baselines. Importantly, our method remains generation-free at inference time, making it more efficient in practice than existing methods.

\paragraph{Contributions.}
First, we introduce a new binary matching $m$ for an input $x$, context $t$, and answer candidate $y$,
define a match-conditioned posterior over answer candidates (\cref{sec:matching_prob_intro}), and quantify its uncertainty with \renyitwoent (\cref{sec:uncertainty_decomposition}). This approach yields a novel decomposition into a \textit{context-specific term}, which quantifies how broadly the context defines the answer candidate space, and a \textit{multiplicity-specific term}, which quantifies how many answer candidates remain plausible with the observed input. Second, we develop a practical estimator for the proposed uncertainty by employing an MLLM verifier to estimate matching probabilities (\cref{sec:matching_prob}), approximating $p(y\mid t)$ by collecting answer candidates from visual instruction tuning datasets (\cref{sec:p_y_t_dataset_construction}), and training lightweight post-hoc heads to predict uncertainty from $(x,t)$ alone (\cref{sec:training}). We refer to the proposed pipeline as \ours. Third, \ours outperforms existing baselines across open-ended VQA, hallucination detection, and MCQ benchmarks, while we can take advantage of our decomposed components, \eg, interpretable information about different sources of ambiguity (\cref{sec:exp_analysis}).

\section{Related Work}

\textbf{Sampling-based uncertainty for M/LLM.}
Sampling-based methods estimate uncertainty by aggregating multiple outputs, \eg, quantify the output consistency \cite{wang2022self_consistency} or their semantic meaning \cite{farquhar2024semantic_entropy,aichberger2024improving}. In MLLM, the diverse outputs can also be generated by perturbing input pixels \cite{zhang2024vl_uncertainty}. While these methods can estimate meaningful uncertainty quantities, they require multiple generations, which limits their practicality. We estimate uncertainty efficiently without generation.

\textbf{M/LLM-as-uncertainty estimator.}
Instead of generating samples repeatedly, a line of study attempts to let an LLM directly quantify its uncertainty. For example, \citet{kadavath2022p_correct} derives uncertainty by the probability of the ``True'' token when a user asks whether the generated output is correct. As another example, \citet{verbalized_confidence} lets a model predict its uncertainty in a verbalized form, \eg, ``confidence is 8/10'', along with its generated output. These methods empirically show their effectiveness, but they remain tied to model-specific confidence proxies or generation behavior. On the other hand, our decomposed uncertainty formulation reduces the reliance on the model capability.

In \cref{sec:appendix_relwork}, we provide a more comprehensive discussion, including methods that require expensive fine-tuning and methods that decompose uncertainty. In summary, \ours enables efficient uncertainty estimation inference using a lightweight post-hoc module, and our decomposition provides a theoretically motivated perspective that differs from existing uncertainty decompositions.

\section{\oursfull}
\label{sec:decomposition_main}

In this section, we introduce a novel decomposition of multimodal uncertainty into context-specific and multiplicity-specific terms. Our key idea is to introduce a binary variable, matching $m$, which indicates whether an input $x$ and an answer candidate $y$ are semantically compatible under a context $t$. We then derive a match-conditioned posterior and quantify its uncertainty using \renyitwoent. Our formulation has two advantages: it provides a theoretically motivated uncertainty decomposition, and it can be applied to open-ended settings where the answers cannot be explicitly enumerated.

\subsection{Matching probability to posterior}
\label{sec:matching_prob_intro}

Let $x$ denote the input, $y$ the answer candidate, and $t$ the context that specifies the space of $y$, such as a task description or a prompt. $x, y$ and $t$ can be continuous. In this paper, we assume the simplest scenario, where $x$ is an image, $t$ is a text question or instruction, and $y$ is an answer candidate.

For a given context $t$, let $p(y \mid t)$ denote a context-conditioned distribution over answer candidates. For example, if $t$ is the question ``What is this animal?'', $y$ may be ``a cat'', ``a dog'', or ``a calico cat''.
Note that we define $y$ as an answer candidate of the given context $t$, therefore, it is not yet tied to a particular input $x$ (\ie, $p(y \mid x, t) = p(y \mid t)$). To connect $y$ with an input $x$, we introduce a binary variable $m$, where $m=1$ indicates that candidate $y$ is a plausible answer for input $x$ and context $t$. For example, let $x$ be an image of a cat and $t$ be the question ``What is this animal?''. The candidate $y=$``a cat'' is matched ($m=1$), while $y=$``a dog'' is not ($m=0$).

Our goal is to quantify the uncertainty over plausible answers for a given input $x$ and context $t$. We consider the distribution of answer candidates conditioned on $m=1$, \ie, $\pi_y := p(y \mid x, t, m=1)$:
\begin{align}
\begin{split}
\label{eq:softmax_bayes}
\pi_y = \frac{p(m=1 \mid x, t, y) p(y \mid x, t)}{\int p(m=1 \mid x, t, y') p(y' \mid x, t) d y'} = \frac{p(m=1 \mid x, t, y) p(y \mid t)}{\int p(m=1 \mid x, t, y') p(y' \mid t) d y'},
\end{split}
\end{align}
The equation holds because we assume that a candidate $y$ is drawn using $t$ alone without relating it to a particular $x$. Therefore, before conditioning on $m=1$, the candidate distribution satisfies $p(y \mid t) = p(y \mid x, t)$.
\cref{eq:softmax_bayes} represents the match-conditioned posterior $\pi_y$ with two quantities: (1) the matching probability $p(m=1 \mid x, t, y)$, which measures whether candidate $y$ is a plausible answer for the given $x$ and $t$, and (2) the answer candidate distribution of the given $t$, \ie, $p(y \mid t)$.

Our match-conditioned posterior formulation has two practical advantages. First, given $p(y\mid t)$, \cref{eq:softmax_bayes} enables the estimation of the posterior $\pi_y$ by answer verification (\eg, asking whether the answer is correct \cite{kadavath2022p_correct}) over all answer candidates. Therefore, our formulation does not need to estimate the answer distribution through autoregressive generation, which is sensitive to generation prompts, sampling hyperparameters, and decoding algorithms, and may produce semantically equivalent paraphrases.
Second, $p(y \mid t)$ does not depend on the specific input $x$ or the model. We can approximate it directly from answers to similar questions in an existing image-question-answer dataset. This approximated $p(y \mid t)$ can be reused across inputs and models.

\subsection{Uncertainty Decomposition}
\label{sec:uncertainty_decomposition}

Our goal is to quantify the uncertainty of the match-conditioned posterior, \cref{eq:softmax_bayes}.
If $y$ is discrete and its distribution is explicitly given (\ie, in closed-set settings), we can easily compute its information entropy by 
$-\sum_y \pi_y \log \pi_y$.
For example, in classification, this sum is taken over the target classes.
However, in open-ended settings, 
the answer space is
not explicitly enumeratable. Also, if we model the answer space as continuous, we need another form of entropy defined for continuous space.

To mitigate this problem, we use a generalized notion of entropy, specifically \renyitwoent, defined by $H_2 (p(x)) := - \log \int_x p(x)^2 dx$. \renyient is a generalized version of entropy, which is defined by $H_\alpha (p(x)) := \lim_{\gamma \to \alpha}\frac{1}{1 - \gamma} \log \int p(x)^\gamma dx$; if $\alpha=1$, it is the same as Shannon's entropy in discrete cases and differential entropy in continuous cases. The case of $\alpha=2$, also known as collision entropy or Simpson diversity index, is particularly suitable in our case because it quantifies how concentrated the distribution is. We provide a more detailed discussion in \cref{sec:appendix_differential_entropy}.

Let $Z_1(x, t) := \mathbb E_{y' \sim p(\cdot \mid t)} \big[ p(m=1 \mid x, t, y') \big]$ and $Z_2(x, t) := \mathbb E_{y' \sim p(\cdot \mid t)} \big[ p(m=1 \mid x, t, y')^2 \big]$. We assume that $p(y \mid t)$ is a uniform distribution over a restricted region $\Omega_t$, \ie, $p(y \mid t) = \frac{1}{\int_{\Omega_t} dy} := \frac{1}{K_t}$. Then \renyitwoent of \cref{eq:softmax_bayes} can be computed as follows:
\begin{align}
\begin{split}
\label{eq:decomposed_uncertainty}
u(x, t) &:= - \log \int \pi_y^2 dy = - \log \int \frac{p(m=1 \mid x, t, y)^2 p(y \mid t)^2}{Z_1(x,t)^2}dy = -\log \left[ \frac{1}{K_t}\frac{Z_2(x,t)}{Z_1(x,t)^2} \right]\\
&= \underbrace{\log K_t}_{:=u_t(t)} + \underbrace{2 \log Z_1(x,t) - \log Z_2 (x,t)}_{:=u_{x \mid t}(x \mid t)},
\end{split}
\end{align}
where the more detailed derivation is in \cref{sec:appendix_derivation_renyi}. In \cref{eq:decomposed_uncertainty}, the uncertainty is decomposed by (1) the context-specific term $u_t(t)$ and (2) the multiplicity-specific term $u_{x \mid t} (x \mid t)$. More specifically, $u_t(t)$ captures how broad the plausible answer space for the given context $p(y \mid t)$ is:
\begin{equation}
\label{eq:u_t}
u_t(t) = \log K_t = -\log p(y \mid t) \underset{\text{(If $p(y \mid t)$ is uniform)}}{=} -\log \mathbb E_{Y \sim p(\cdot \mid t)}[p(y\mid t)] = H_2(p(y \mid t)).
\end{equation}
In other words, if the context $t$ is ambiguous to answer, we naturally have a high uncertainty. For example, the context ``describe this image'' ($K_t$ is almost infinitely large) has a higher uncertainty than the context ``Is there a cat in this image? Answer yes or no'' ($K_t$ is 2). 

When we take a look into the multiplicity-specific term $u_{x \mid t}(x \mid t)$, we have the following relationship:
\begin{equation}
\label{eq:u_xt}
u_{x \mid t}(x \mid t) = 2\log Z_1(x,t) - \log Z_2 (x,t) = -\log\left(\frac{\text{Var}_{Y \sim p(\cdot \mid t)}[p(m=1 \mid x, t, Y)]}{\{\mathbb E_{Y \sim p(\cdot \mid t)} [p(m=1 \mid x, t, Y)]\}^2} + 1\right),
\end{equation}
which implies that $u_{x \mid t}(x \mid t)$ can be represented in a form related to coefficient of variation (or relative standard deviation) of the matching probability over the distribution $p(y \mid t)$. In other words, if the matching probabilities $p(m=1 \mid x,t,y)$ are very similar for all the possible $y$ (\ie, how many $y$ can be matched to $x$ with the given $t$), then we have a high uncertainty value. We provide a more detailed discussion related to $u_t$ and $u_{x \mid t}$ in \cref{sec:appendix_derivation_renyi}.
We also show that we can have a similar formulation with \cref{eq:decomposed_uncertainty} even with discrete entropy. Details are provided in \cref{sec:appendix_shannon_entropy_derivation_taylor}.

\begin{figure}[t]
    \centering
    \includegraphics[width=\linewidth]{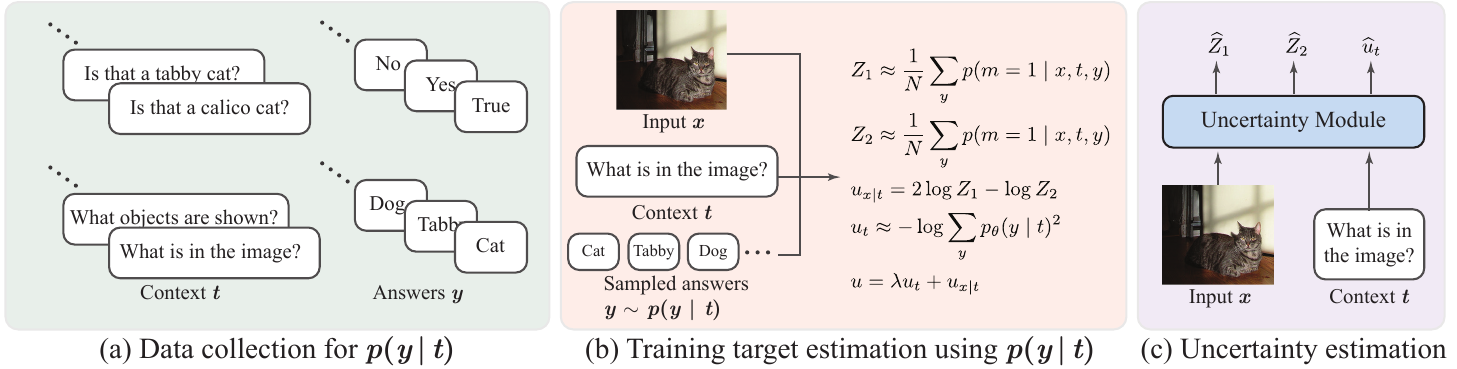}
    \caption{\small {\bf Overview of the proposed \ours framework.} (a) We construct an answer candidate distribution $p(y\mid t)$ based on text clustering on the Cambrian dataset \cite{tong2024cambrian} (\cref{sec:p_y_t_dataset_construction}). (b) Using the MLLM-based matching probability estimator (\cref{sec:matching_prob}), we estimate the uncertainty targets from the constructed dataset. (c) We train a lightweight uncertainty module using the constructed dataset and the estimated targets (\cref{sec:training}).}
    \label{fig:method_overview}
\end{figure}

\section{Practical Uncertainty Estimation}
\label{sec:method}

Uncertainty estimation with our formulation, \cref{eq:decomposed_uncertainty}, requires two quantities: the matching probability $p(m=1 \mid x,t,y)$ and the context-conditioned answer candidate distribution $p(y \mid t)$. For the matching probability, we adopt an MLLM-as-verifier strategy. Estimating $p(y \mid t)$ is more challenging. We construct a answer candidate set that approximates $p(y \mid t)$, without autoregressive candidate generation, avoiding its generation-related sensitivity and cost. Using these two ingredients, we train lightweight post-hoc uncertainty modules that predict our uncertainty formulation from an input $x$ and a context $t$. Once the module is trained, the uncertainty can be estimated from $x$ and $t$ alone without requiring any generation process or sampling process from an MLLM, which makes the overall inference efficient. The overview of our framework \ours is illustrated in \cref{fig:method_overview}.

\subsection{Verification as matching probability}
\label{sec:matching_prob}

Our decomposition requires a matching probability function $p(m=1 \mid x, t, y)$. We can employ a parameterized matching probability module, which is a binary classifier that takes a $(x, t, y)$ triplet as input. However, we found that training a reliable matching probability module is difficult and does not generalize well to arbitrary open-ended settings. More detailed discussions are in \cref{sec:appendix_why_no_train_matching_probability}.

Instead of training a separate matching probability module, we employ an MLLM as a verifier of matching. For example, we let an MLLM take ``\texttt{Image:\{x\},} \texttt{Question:\{t\},} \texttt{Proposed Answer:\{y\}.} \texttt{Is this answer correct?} \texttt{Answer Yes or No.}'', then compute the probability of \texttt{``Yes''} (the full prompt can be found in \cref{sec:appendix_matching_verification_template}). This simple technique is known to be effective in estimating self-correctness despite its simplicity \cite{kadavath2022p_correct,lin2024evaluating}. We additionally employ affine transform parameters $(\beta, \gamma)$ to calibrate the matching probability as follows:
\begin{equation}
\label{eq:verify_matching_probability}
p(m =1 \mid x, t, y) = \text{sigmoid} \left(\beta \log \frac{p(\text{``Yes''} \mid x, t, y)}{p(\text{``No''} \mid x, t, y)} + \gamma\right),
\end{equation}
If $\beta=1, \gamma=0$, \cref{eq:verify_matching_probability} becomes $\frac{p(\text{``Yes''} \mid x, t, y)}{p(\text{``Yes''} \mid x, t, y)+p(\text{``No''} \mid x, t, y)}$. In our experiments, we calibrate $\beta$ and $\gamma$ from a validation dataset. See more details in \cref{sec:appendix_beta_gamma_calibration}.

\subsection{Approximating the context-conditioned answer candidate distribution}
\label{sec:p_y_t_dataset_construction}

We also require $p(y \mid t)$ for our uncertainty estimates. One possible approach is to approximate it using a language model, \ie, $p(y \mid t) \approx p_\theta(y \mid t)$.
However, this approach is sensitive to generation hyperparameters, such as prompts, sampling hyperparameters and decoding algorithms. Also, it is computationally expensive as we observed in our experiment. Instead, we efficiently approximate $p(y\mid t)$ from existing image-question-answer datasets without autoregressive generation.

We first collect all question texts and their corresponding answers from the Cambrian dataset \cite{tong2024cambrian}. Then, we perform a simple greedy clustering on the questions. After clustering, we treat all the answers from the same cluster as the answer candidates of the questions in the cluster. For example, assume that we have a question ``what is in this image?'' and its answer ``a cat'', and another question-answer pair ``what is in this photo?'' and ``a bird'' in a cluster. In this case, both ``a cat'' and ``a bird'' are treated as answer candidates of the question ``what is in this image?''.
Compared to sampling-based methods \cite{farquhar2024semantic_entropy} constructing a finite answer set by repeated generation, we construct the set by reusing answers in existing datasets. This avoids dependence on generation prompts and decoding hyperparameters, while grounding candidate construction in observed question–answer pairs.
We describe the details of the collection procedure in \cref{sec:appendix_dataset_collection}.

\subsection{Training objective}
\label{sec:training}

Now, we approximate the quantities $Z_1$, $Z_2$ (\cref{eq:u_xt}), $u_t$ (\cref{eq:u_t}), and $u$ (\cref{eq:decomposed_uncertainty}) defined in \cref{sec:uncertainty_decomposition} using $p(m=1 \mid x, t, y)$ in \cref{sec:matching_prob} and $p(y \mid t)$ in \cref{sec:p_y_t_dataset_construction}.

First, we approximate $u_{x \mid t} = 2 \log Z_1(x, t) - \log Z_2(x, t)$ by Monte Carlo approximation using a finite answer candidate set $\mathcal Y_t$ that approximates $p(y \mid t)$ as follows:
\begin{equation}
\label{eq:u_x_approximation}
\widehat Z_1(x, t) = \frac{1}{|\mathcal Y_t|} \sum_{y' \in \mathcal Y_t} p(m=1 \mid x, t, y') \quad \widehat Z_2(x, t) = \frac{1}{|\mathcal Y_t|} \sum_{y' \in \mathcal Y_t} p(m=1 \mid x, t, y')^2.
\end{equation}
Estimating $u_t(t)$ is more challenging because we cannot access the true $p(y \mid t) = \frac{1}{K_t}$ distribution, hence $K_t$ is unknown. Instead, we employ a practical approximation using a tractable density estimator $p_\theta(y \mid t)$, \ie, an autoregressive LM, based on \cref{eq:u_t} as follows:
\begin{equation}
\label{eq:u_t_approximation}
u_t(t) = H_2 (p(y \mid t)) \approx \widehat u_t(t) = -\log \sum_{y' \in \mathcal Y_t} p_\theta(y'\mid t)^2,
\end{equation}
where $p_\theta(y'\mid t)$ can be computed using the autoregressive nature of LMs. Since this surrogate breaks the uniform assumption of $p(y \mid t)$, \cref{eq:u_t_approximation} is not exactly the same as the original $u_t(t)$. To control the effect of the approximation error, we add a control hyperparameter $\lambda$ as follows:
\begin{equation}
\label{eq:decomposed_uncertainty_approximation}
\widehat u(x, t) = \lambda \widehat u_t(t) + 2 \log \widehat Z_1(x, t) - \log \widehat Z_2(x, t).
\end{equation}
Using the collected data samples described in \cref{sec:p_y_t_dataset_construction}, we optimize a light-weight Transformer head that estimates $\widehat Z_1(x, t), \widehat Z_2(x, t)$ and $\widehat u_t(t)$ using a regression loss. After training the module, we additionally employ a light-weight adaptation to the attention pooling layers and the last linear layers of the uncertainty module on a small number of data samples related to the target benchmark.
More details of the model architecture and the objective functions are described in \cref{sec:appendix_architecture}.

\section{Experiments}

In this section, we evaluate \ours for its capability to
improve MLLM uncertainty estimation. We first evaluate \ours on open-ended multimodal understanding, unanswerability/hallucination detection, and multiple-choice question (MCQ) settings (\cref{sec:exp_main}). Next, we examine the effectiveness of our proposed decomposition formulation by quantitative and qualitative studies (\cref{sec:exp_analysis}).

\subsection{Experimental setup}
\label{sec:exp_settings}
\textbf{Benchmarks.}
We evaluate \ours on open-ended VQA benchmarks, including VQA v2 \cite{vqav2}, VizWiz \cite{vizwiz}, and OK-VQA \cite{okvqa}. For VQA v2, we randomly subsample 10k samples from the dataset to reduce the inference cost of sampling or generation-based methods (the original VQA v2 consists of 214k samples).
We additionally consider two complementary reliability tasks: unanswerability detection on VizWiz and hallucination detection on HallusionBench \cite{hallusionbench}.
Finally, we evaluate \ours on MCQ VQA benchmarks, including MMMU (test split) \cite{mmmu}, MMMU Pro \cite{mmmu_pro}, and MMStar \cite{mmstar}. We describe more details of each benchmark in \cref{sec:appendix_evaluation_benchmark}.

\textbf{Evaluation metrics.}
We report standard uncertainty estimation metrics, such as AUROC, AUPR, and FPR95. Following common practice, we treat uncertainty estimation as a binary classification problem where the uncertainty score is used to distinguish correct predictions from incorrect ones. For open-ended VQA whose evaluation is based on a soft VQA score, we use a threshold to make the prediction binary. We describe the details of our evaluation metrics in \cref{sec:appendix_evaluation_metric}.

\textbf{Implementation details.}
Using the data collection process described in \cref{sec:p_y_t_dataset_construction}, we collect 200,000 image-question pairs, where each pair is associated with 32 plausible answers. We split this dataset into 199k training samples and 1k validation samples. Model selection is performed on the 1k validation split, and the hyperparameter search is performed on the MMMU validation split (900 samples). 
In our hyperparameter study, we set $\beta=0.2$ and $\gamma=-0.1$ for \cref{eq:verify_matching_probability} across all backbones. For \cref{eq:decomposed_uncertainty_approximation}, we use $\lambda=0.25$ for all backbones except Qwen3VL-4B-Instruct, for which $\lambda=1.0$ performs best.
For adaptation, we use 900 samples of the MMMU validation split \cite{mmmu} and 10k randomly selected samples from VQA v2 \cite{vqav2}, without overlap with our selected 10k test split.

We evaluate four MLLM backbones: Qwen3VL-2B-Instruct, Qwen3VL-4B-Instruct, Qwen3VL-8B-Instruct \cite{bai2025qwen3}, and InternVL3.5-1B \cite{wang2025internvl3_5}. For each backbone, we use the selected backbone for matching probability estimator (\cref{eq:verify_matching_probability}), and train a separate uncertainty module while keeping the original MLLM weight fixed.
The full implementation details are provided in \cref{sec:appendix_implementation_details}.

\textbf{Comparison methods.}
We compare \ours against four groups of methods. (1) \textit{Generation-based} that generate an answer $y$ for a given input $x$ and context $t$, and then estimate confidence from the generated output. \textbf{Negative log-likelihood (NLL)} measures the likelihood of the generated answer under $p(y \mid x, t)$, \textbf{P(Correct)} \cite{kadavath2022p_correct} estimates confidence from the token probability of ``Yes'' when the model is asked a follow-up question, \eg, ``Is this answer correct?''. \textbf{Verbalized confidence} \cite{verbalized_confidence} asks the model to produce its confidence in text, \eg, ``confidence is 6/10''. (2) \textit{Sampling-based} that generate multiple answers for the same input and estimate uncertainty from their consistency. We consider \textbf{consistency entropy} and \textbf{consistency maximum probability}, both computed from the empirical answer distribution induced by repeated sampling \cite{wang2022self_consistency}. (3) \textit{Perturbation-based}: we compare against \textbf{VL-Uncertainty} \cite{zhang2024vl_uncertainty}, which derives MLLM uncertainty from consistency of generated outputs with perturbed images. (4) \textit{Candidate-based baselines}: for MCQ VQA, we additionally report \textbf{candidate entropy} and \textbf{candidate maximum probability}, computed over the explicit answer options. More details of the methods are provided in \cref{sec:appendix_comparison_methods}.

\begin{table}[t!]
    \centering
    \footnotesize
    \caption{\small {\bf Main comparison results.} We report AUROC (AUC), AUPRC (AP), and FPR@95 of uncertainty estimation methods on three settings: open-ended multimodal understanding, error detection, and MCQ VQA. We report the average score for each setting; the full results are in \cref{sec:appendix_full_results}. ``GF'' denotes generation-free.}
    \label{tab:main}
\setlength{\tabcolsep}{3pt}
\begin{tabular}{@{}llllllllllllll@{}}
\toprule
 &  & \multicolumn{3}{l}{Qwen3VL-2B-Instruct} & \multicolumn{3}{l}{Qwen3VL-4B-Instruct} & \multicolumn{3}{l}{Qwen3VL-8B-Instruct} & \multicolumn{3}{l}{InternVL3.5-1B} \\ \midrule
 & GF & AUC $\uparrow$ & AP $\uparrow$ & FPR $\downarrow$ & AUC $\uparrow$ & AP $\uparrow$ & FPR $\downarrow$ & AUC $\uparrow$ & AP $\uparrow$ & FPR $\downarrow$ & AUC $\uparrow$ & AP $\uparrow$ & FPR $\downarrow$ \\ \midrule
\multicolumn{14}{c}{Open-ended multimodal understanding benchmarks (Average)} \\ \midrule
Consist. (Ent) \cite{wang2022self_consistency} & \nomark & .624 & .593 & .918 & .661 & .464 & .899 & .657 & .548 & .904 & .495 & .411 & .959 \\
Consist. (Max) \cite{wang2022self_consistency} & \nomark & .623 & .593 & .914 & .660 & .464 & .899 & .655 & .547 & .904 & .495 & .411 & .959 \\
NLL & \nomark & .569 & .545 & .963 & .491 & .357 & .974 & .475 & .417 & .970 & .576 & .433 & .949 \\
Verbalized Conf \cite{verbalized_confidence} & \nomark & .591 & .592 & .842 & .562 & .378 & .896 & .642 & .598 & .787 & .639 & .546 & .840 \\
P(Correct) \cite{kadavath2022p_correct} & \nomark & .685 & .648 & .863 & .712 & .536 & .837 & .714 & .652 & .766 & .667 & .532 & .897 \\
VL Uncertainty \cite{zhang2024vl_uncertainty} & \nomark & .650 & .646 & .852 & .654 & .605 & .829 & .654 & .665 & .829 & .637 & .535 & .869 \\
\rowcolor{gray!20}
\ours (Ours) & \yesmark & \textbf{.785} & \textbf{.784} & \textbf{.707} & \textbf{.784} & \textbf{.631} & \textbf{.717} & \textbf{.799} & \textbf{.728} & \textbf{.701} & \textbf{.734} & \textbf{.647} & \textbf{.771} \\
\midrule
\multicolumn{14}{c}{Unanswerability/hallucination detection benchmarks (Average)} \\ \midrule
Consist. (Ent) & \nomark & .586 & .519 & .886 & .599 & .438 & .877 & .614 & .442 & .886 & .493 & .441 & .926 \\
Consist. (Max) & \nomark & .584 & .519 & .892 & .598 & .434 & .904 & .610 & .441 & .886 & .493 & .441 & .926 \\
NLL & \nomark & .506 & .465 & .961 & .547 & .377 & .943 & .499 & .353 & .962 & .514 & .432 & .950 \\
Verbalized Conf & \nomark & .599 & .567 & .953 & .548 & .397 & .930 & .612 & .503 & .911 & .615 & .560 & .954 \\
P(Correct) & \nomark & .672 & .619 & .817 & .648 & .474 & .800 & .635 & .467 & .844 & .649 & .553 & .812 \\
VL Uncertainty & \nomark & .534 & .584 & .955 & .522 & .380 & .979 & .522 & .376 & .975 & .561 & .526 & .884 \\
\rowcolor{gray!20}
\ours (Ours) & \yesmark & \textbf{.751} & \textbf{.693} & \textbf{.710} & \textbf{.705} & \textbf{.540} & \textbf{.761} & \textbf{.701} & \textbf{.572} & \textbf{.765} & \textbf{.732} & \textbf{.649} & \textbf{.767} \\ \midrule
\multicolumn{14}{c}{Multiple choice question (MCQ) visual question answering (VQA) benchmarks (Average)} \\ \midrule
Candidate (Ent) & \yesmark & .719 & .801 & .799 & .680 & .676 & .861 & .675 & .646 & .846 & .583 & .657 & .921 \\
Candidate (Max) & \yesmark & .708 & .793 & .812 & .673 & .664 & .880 & .669 & .635 & .867 & .575 & .651 & .928 \\
Consist. (Ent) & \nomark & .775 & .899 & .562 & .709 & .813 & \textbf{.638} & .714 & .791 & .658 & \textbf{.680} & \textbf{.804} & \textbf{.767} \\
Consist. (Max) & \nomark & .766 & .893 & .586 & .706 & .807 & .678 & .709 & .781 & .688 & .674 & .798 & .781 \\
NLL & \nomark & .191 & .556 & .996 & .247 & .446 & .996 & .301 & .452 & .980 & .357 & .558 & .994 \\
Verbalized Conf & \nomark & .617 & .773 & .933 & .540 & .676 & .911 & .560 & .640 & .899 & .531 & .695 & .892 \\
P(Correct) & \nomark & .511 & .688 & .970 & .729 & .776 & .763 & .733 & .771 & .724 & .589 & .716 & .911 \\
VL Uncertainty & \nomark & .802 & .909 & .482 & .721 & .812 & .660 & .722 & .789 & .687 & .587 & .715 & .902 \\
\rowcolor{gray!20}
\ours (Ours) & \yesmark & \textbf{.842} & \textbf{.929} & \textbf{.438} & \textbf{.782} & \textbf{.843} & \underline{.655} & \textbf{.783} & \textbf{.825} & \textbf{.654} & \underline{.660} & \underline{.772} & \underline{.851} \\ \bottomrule
\end{tabular}
\vspace{-1em}
\end{table}

\subsection{Main results}
\label{sec:exp_main}
\vspace{-.25em}
\textbf{Open-ended VQA.}
We first evaluate \ours on open-ended VQA benchmarks, including VQA v2, VizWiz, and OK-VQA. The first part of \cref{tab:main} and \cref{tab:openset_vqa} show that \ours consistently outperforms existing baselines across all three benchmarks and all backbones. For example, on the Qwen3VL-2B backbone, \ours improves the average score by +0.100 AUROC over the strongest baseline, P(Correct).
Importantly, \ours is generation-free at inference time, yet substantially outperforms MLLM-specific uncertainty estimation methods requiring repeated generation, \eg, VL Uncertainty \cite{zhang2024vl_uncertainty}, with a large gap (\eg, +0.135 average AUROC). 

\textbf{Error detection.}
We next evaluate whether the proposed uncertainty estimates transfer to related reliability tasks, such as unanswerability detection on VizWiz and hallucination detection on HallusionBench. They are complementary to open-ended VQA because they require uncertainty estimates to identify predictions that are unsupported or unreliable, rather than rank answer confidence.
As shown in the second part of \cref{tab:main}, \ours achieves the best average performance across all backbones. The full results in \cref{tab:unanswer_hallu_detection_results} show that \ours achieves either the best or second-best performance on all sub-benchmarks. This supports that our decomposed uncertainty formulation can be extended to more general uncertainty use cases.

\textbf{MCQ VQA.}
Finally, we evaluate \ours on MCQ VQAs, including MMMU, MMMU Pro, and MMStar. Although our formulation is designed for open-ended settings rather than closed-set MCQ tasks, \cref{tab:main} and \cref{tab:mcq_vqa} show that \ours remains highly competitive in this structured setting. In most cases, \ours outperforms the comparison methods; for some cases, such as InternVL3.5-1B, \ours achieves the second-best performance after consistency-based baselines. We presume that this is because MCQ benchmarks provide an explicit candidate set, making repeated sampling particularly effective for estimating answer instability by reducing the search space. Nevertheless, \ours remains competitive without repeated answer generation, suggesting that the proposed decomposed uncertainty captures semantic ambiguity that transfers beyond open-ended settings.

\subsection{Analysis}
\label{sec:exp_analysis}

\begin{table}[t!]
\vspace{-.5em}
\small
\centering
\setlength{\tabcolsep}{4pt}
\caption{\small {\bf Impact of the proposed \ours uncertainty decomposition.} We compare \ours trained with various combinations of uncertainty components in AUROC scores on each benchmark. Qwen3VL-8B-Instruct is used for the backbone, and the MMMU validation split is used for the actual hyperparameter selection.}
\label{tab:loss_ablation}
\begin{tabular}{@{}llcccccccc@{}}
\toprule
$u_t$ & $u_{x \mid t}$ & MMMU val & VQA v2 & VizWiz & OK-VQA & MMMU Test & MMMU Pro & MMStar & Avg \\ \midrule
\yesmark & \nomark & .751 & .691 & .691 & .625 & .716 & .649 & .677 & .686 \\
\nomark & \yesmark & .837 & .834 & .734 & .735 & .753 & .707 & \textbf{.857} & .780 \\
\rowcolor{gray!20}
\yesmark & \yesmark & \textbf{.842} & \textbf{.836} & \textbf{.815} & \textbf{.746} & \textbf{.770} & \textbf{.726} & .853 & \textbf{.798} \\
\bottomrule
\end{tabular}
\vspace{-1em}
\end{table}

\begin{table}[t!]
\small
\centering
\setlength{\tabcolsep}{4pt}
\caption{\small \textbf{Sensitivity to the choice of verifier.}
We compare various verifier models for \ours with the Qwen3VL-2B backbone.
VW and HB denote VizWiz and HallusionBench; (U) denotes unanswerable detection.}
\label{tab:verifier_ablation}
\begin{tabular}{@{}lccccccccc@{}}
\toprule
Verifier & VQA v2 & VW & OK-VQA & VW (U) & HB & MMMU & MMMU Pro & MMStar & Avg \\ \midrule
Qwen3VL-8B & .825 & .782 & .739 & .779 & .672 & .874 & .815 & \textbf{.830} & .789 \\
InternVL3.5-1B & .815 & .741 & .716 & .756 & .695 & .873 & .827 & .815 & .780 \\
\rowcolor{gray!20}
\textbf{Qwen3VL-2B} & \textbf{.833} & \textbf{.783} & \textbf{.740} & \textbf{.788} & \textbf{.715} & \textbf{.875} & \textbf{.834} & .819 & \textbf{.798} \\ \bottomrule
\end{tabular}
\vspace{-.5em}
\end{table}

\begin{table}[t]
\small
\centering
\setlength{\tabcolsep}{4pt}
\caption{\small \textbf{Effect of candidate construction.}
We compare candidates collected from the Cambrian dataset with candidates generated by LLM. $N$ denotes the number of questions. 
Other details are the same as \cref{tab:verifier_ablation}.
}
\label{tab:candidate_ablation}
\begin{tabular}{@{}lcccccccccc@{}}
\toprule
Source & $N$ & VQA v2 & VW & OK-VQA & VW (U) & HB & MMMU & MMMU Pro & MMStar & Avg \\ \midrule
Cambrian & 3k & \textbf{.835} & .741 & .727 & .751 & .707 & \textbf{.879} & .831 & .813 & .786 \\
LLM & 3k & .829 & .764 & .728 & .770 & \textbf{.715} & .878 & .834 & \textbf{.822} & .792 \\
\rowcolor{gray!20}
\textbf{Cambrian} & 200k & .833 & \textbf{.783} & \textbf{.740} & \textbf{.788} & \textbf{.715} & .875 & \textbf{.834} & .819 & \textbf{.798} \\ \bottomrule
\end{tabular}
\vspace{-.5em}
\end{table}

\begin{table}[t!]
\small
\centering
\setlength{\tabcolsep}{4pt}
\caption{\small \textbf{Impact of adaptation dataset.} The adaptation results with various datasets are shown. For Cambrian, the adaptation and the pre-training datasets are from the same source.
Other details are the same as \cref{tab:verifier_ablation}.
}
\label{tab:adaptation_comparison}
\begin{tabular}{@{}cccccccccc@{}}
\toprule
Adaptation dataset & VQA v2 & VW & OK-VQA & VW (U) & HB & MMMU & MMMU Pro & MMStar & Avg \\ \midrule
No adaptation & .580 & .648 & .609 & .753 & .572 & .546 & .513 & .518 & .592 \\
Cambrian & .638 & .714 & .663 & .734 & .684 & .641 & .642 & .762 & .685 \\
\rowcolor{gray!20}
MMMU + VQA v2 & \textbf{.833} & \textbf{.783} & \textbf{.740} & \textbf{.788} & \textbf{.715} & \textbf{.875} & \textbf{.834} & \textbf{.819} & \textbf{.798} \\ \bottomrule
\end{tabular}
\vspace{-.5em}
\end{table}

\begin{table}[t!]
\small
\centering
\setlength{\tabcolsep}{4pt}
\caption{\small \textbf{Comparison with direct post-hoc training methods.} We compare \ours after adaptation with uncertainty modules directly trained using BCE or ranking loss on the same adaptation data. All methods use the same frozen MLLM features (Qwen3VL-2B-Instruct) and uncertainty-module architecture. ``All layers'' updates the entire uncertainty module; ``Last layer'' updates only the attention pooling layer and last linear layer.}
\label{tab:direct_posthoc_comparison}
\resizebox{\linewidth}{!}{
\begin{tabular}{@{}clccccccccc@{}}
\toprule
Method & Adaptation & VQA v2 & VW & OK-VQA & VW (U) & HB & MMMU & MMMU Pro & MMStar & Avg \\ \midrule
Direct (BCE) & All layers & .825 & .772 & .706 & .773 & .595 & .859 & \textbf{.837} & .805 & .772 \\
Direct (Ranking) & All layers & .791 & .763 & .694 & \textbf{.793} & .680 & .852 & .834 & .791 & .775 \\
\rowcolor{gray!20}
\textbf{CoMet + Adapt.} & Last layer & \textbf{.833} & \textbf{.783} & \textbf{.740} & .788 & \textbf{.715} & \textbf{.875} & .834 & \textbf{.819} & \textbf{.798} \\ \bottomrule
\end{tabular}
}
\vspace{-.5em}
\end{table}

\textbf{Effect of decomposed uncertainty targets.}
\cref{tab:loss_ablation} shows that training a model with only $u_t$ or $u_{x \mid t}$ performs consistently worse than using both, \eg, 0.798 average AUROC with both vs. 0.780 or 0.686 with each term alone. This shows that the two terms capture complementary uncertainty signals.
In particular, $u_t$ can be effective when dataset-level context priors are informative, while $u_{x\mid t}$ becomes important when the visual input is necessary to resolve the ambiguity (\eg, VQA v2 \cite{vqav2}).
In \cref{sec:appendix_additional_loss_ablation}, we additionally compare \ours with unified entropy targets without decomposition; \ours achieves 0.798 average AUROC, while the unified entropy target achieves 0.729.
These results support the effectiveness of our multimodal uncertainty decomposition.

\textbf{Sensitivity to the MLLM-as-verifier.} We examine how the choice of verifier affects \ours while keeping the Qwen3VL-2B backbone and all other settings fixed.
We observe that in their individual matching decisions, the verifiers differ substantially: InternVL3.5-1B and Qwen3VL-8B disagree with Qwen3VL-2B on 35.0\% and 19.3\% of collected candidates from \cref{sec:p_y_t_dataset_construction}.
Despite these differences, Qwen3VL-2B, Qwen3VL-8B, and InternVL3.5-1B verifiers achieve comparable average AUROCs of 0.798, 0.789, and 0.780, respectively (\cref{tab:verifier_ablation}).
Aggregating matching probabilities over $p(y \mid t)$ may explain this limited sensitivity. We provide more detailed discussion in \cref{sec:appendix_additional_analysis}.

\textbf{Effect of candidate construction.}
We compare our dataset-based candidate construction strategy introduced in \cref{sec:p_y_t_dataset_construction} with generation-based construction. We generate 28 candidates for each question using Qwen3-8B \cite{yang2025qwen3}; note that we need to design a dedicated candidate-generation prompt and in-context examples, where we use our dataset-based candidates as the in-context samples -- Details are in \cref{sec:appendix_additional_analysis}.
Using the same 3k questions, LM-generated candidates achieve 0.792 average AUROC, compared with 0.786 for dataset-based candidates, showing that \ours remains effective under the alternative strategy (\cref{tab:candidate_ablation}). 
However, LM construction for 3k questions requires 84k generations, requiring 250 L40 GPU-hours. Meanwhile, our dataset-based method only takes fewer than 2.5 L40 GPU-hours for 200k questions.
As shown in \cref{tab:candidate_ablation} and \cref{sec:appendix_additional_analysis}, increasing the number of questions or candidates per question tends to improve average performance.
In summary, our candidate construction strategy provides a more scalable candidate-construction mechanism without carefully designed generation prompts or hyperparameters.

\textbf{Impact of adaptation dataset.}
We examine how the source of adaptation data affects the downstream performance of \ours.
\cref{tab:adaptation_comparison} compares the results without adaptation, with adaptation samples constructed from the pre-training dataset only, and with adaptation on held-out samples from downstream tasks.
Without adaptation, \ours achieves 0.592 average AUROC; Cambrian-only adaptation improves it to 0.685 without using downstream benchmark data, comparable to VL-Uncertainty (0.678).
\cref{sec:appendix_adaptation_ablation} shows more detailed results with various adaptation datasets.

\textbf{Comparison with direct post-hoc predictors.}
Finally, we compare \ours with the uncertainty module trained on the same adaptation dataset. We train uncertainty modules with BCE or the proposed ranking adaptation loss in \cref{eq:adaptation_objective}.
Here, the comparison methods update the entire uncertainty module parameters, while \ours adapts only the pooling and last linear layer.
\ours achieves 0.798 average AUROC, outperforming BCE (0.772) and ranking (0.775) post-hoc modules.
The result supports that the effectiveness of \ours is not explained by adaptation alone.

\textbf{Controlled analysis.}
We examine how $u_t$ and $u_{x\mid t}$ change by altering the input $x$ or the context $t$. As expected, making $x$ more ambiguous increases $u_{x\mid t}$ while leaving $u_t$ fixed. When we intervene in $t$ to allow more answers, $u_t$ increases.
The detailed discussion can be found in \cref{subsec:intervention_appendix}.

\begin{figure}[t!]
    \centering
    \includegraphics[width=\linewidth]{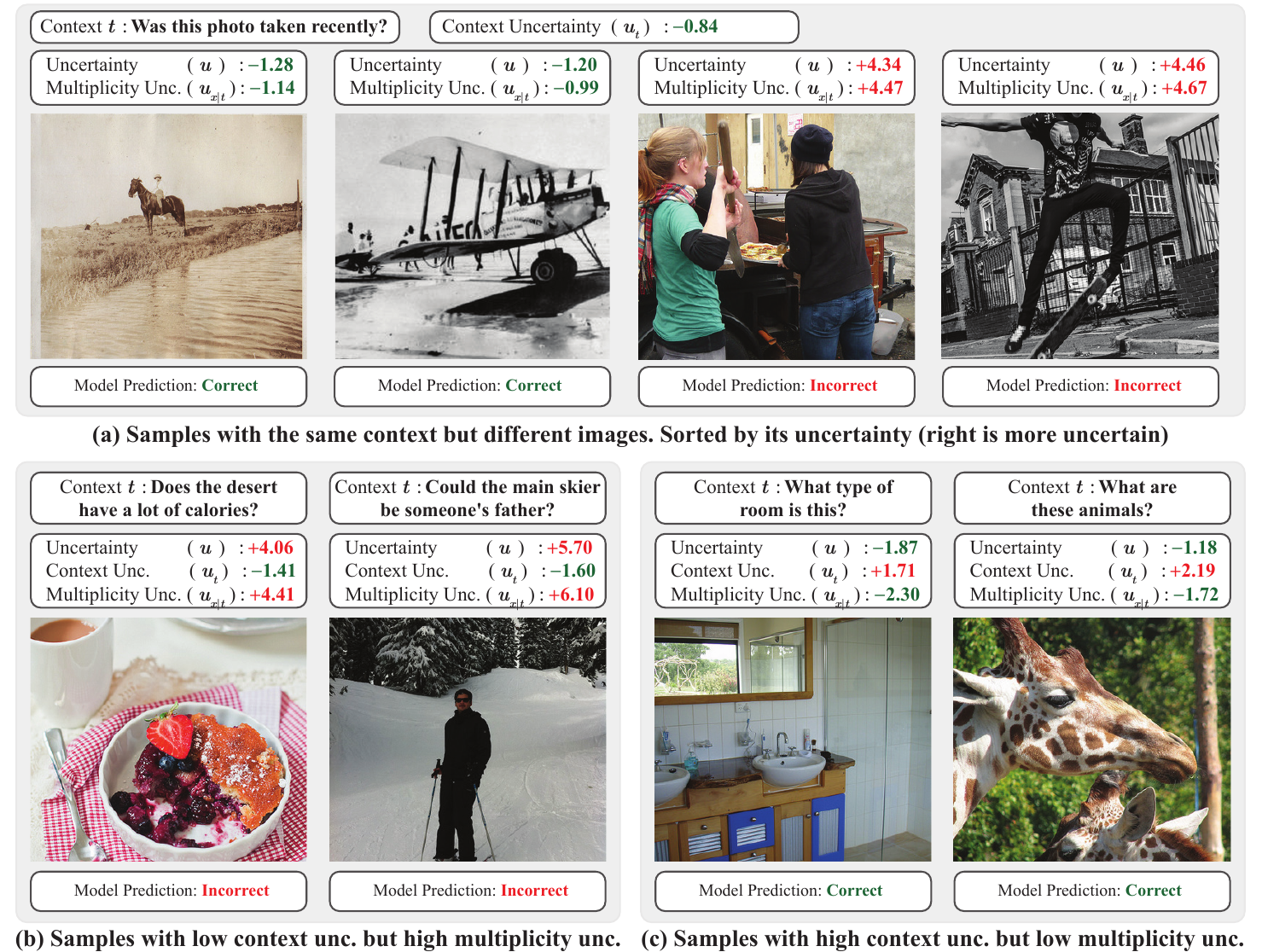}
    \caption{\small {\bf Visualization of uncertainties of various samples.}}
    \label{fig:main_visualization}
\end{figure}

\textbf{Visualization of uncertain samples.}
\cref{fig:main_visualization} visualizes examples with the estimated uncertainty $u$ and its decomposed components $u_t$ and $u_{x \mid t}$. \cref{fig:main_visualization} (a) shows that the uncertainty can vary through $u_{x \mid t}$ even when the context $t$ is fixed; \cref{fig:uncertainty_teaser} shows the complementary case where the image is fixed but the context changes.
Since the question is a yes/no question, it yields low $u_t$. However, when the image does not provide enough visual evidence to determine ``when the photo was taken'', $u_{x \mid t}$ becomes high, as shown in the rightmost examples.
\cref{fig:main_visualization}(b) and (c) show cases where $u_t$ and $u_{x \mid t}$ behave differently. For example, the question ``What are these animals?'' induces a broad answer space, but the image clearly identifies the animals as giraffes, resulting in high $u_t$ but low $u_{x \mid t}$.

We additionally provide more visualizations of the learned uncertainty in \cref{sec:appendix_visualization}.

\textbf{Additional analyses.}
We provide further analyses in the appendix.
\cref{sec:appendix_parameter_study} studies the effect of $\lambda$ (for \cref{eq:decomposed_uncertainty_approximation}) and $(\beta,\gamma)$ (for \cref{eq:verify_matching_probability}).
\cref{sec:appendix_runtime} reports runtime comparisons, where sampling-based baselines \cite{wang2022self_consistency,zhang2024vl_uncertainty} are 2.8 to 5.9 times slower than \ours in the setting. 

\section{Conclusion}

We introduced \ours, a decomposition-based MLLM uncertainty estimation framework that separates context-specific ambiguity from multiplicity-specific uncertainty. By combining an MLLM-as-verifier strategy with lightweight post-hoc uncertainty heads, \ours enables efficient uncertainty estimation in open-ended multimodal settings without autoregressive answer generation or repeated sampling. Experiments across open-ended multimodal understanding, error detection, and MCQ VQA benchmarks show that this decomposition yields more reliable uncertainty estimates than existing baselines while preserving practical inference efficiency.
We hope that our decomposed formulation offers a useful step toward understanding and estimating uncertainty in MLLMs.

\section*{Acknowledgements}
This work was partially supported by the Princeton Francis Robbins Upton Fellowship (SC), NSF Athena \#2112562 (OR), and NSF CAREER \#2145198 (OR). Any opinions, findings, and conclusions or recommendations expressed in this material are those of the authors and do not necessarily reflect the views of the National Science Foundation (NSF).
We are also grateful to Deva Ramanan, Ryan P. Adams, Zhuang Liu, Allison Chen, Salma Abdel Magid, Tyler Zhu, Arnold Caleb Asiimwe, and Song Park for insightful discussions and valuable feedback that helped shape this work.

{\small
\bibliography{bibtex/core,bibtex/sanghyuk,bibtex/uncertainty,bibtex/vision_language}
\bibliographystyle{unsrtnat}
}
\clearpage
\clearpage
\appendix
\crefalias{section}{appendix}
\crefalias{subsection}{appendix}
\crefalias{subsubsection}{appendix}

\numberwithin{equation}{section}
\numberwithin{figure}{section}
\numberwithin{table}{section}

\section*{Appendix}

\section{More Related Work}
\label{sec:appendix_relwork}

\paragraph{Uncertainty estimation.}
Uncertainty estimation has been studied in deep neural networks. Early works in closed-set tasks (\eg, classification) focused on predicting uncertainty scores based on their prediction probability, \eg, entropy \cite{kendall2017uncertainties} or softmax confidence score \cite{hendrycks2016baseline}. These methods have demonstrated effectiveness on calibration \cite{guo2017calibration} or out-of-distributed data detection \cite{hendrycks2016baseline}. However, if the answer space is not explicitly given, these methods are no longer applicable.

\paragraph{Improving uncertainty estimation via fine-tuning.}
Although using M/LLM as uncertainty estimator empirically shows meaningful uncertainty estimation, these methods need additional fine-tuning of the backbone LM \cite{kapoor2024large,xiao2026vl_calibration} to achieve a proper uncertainty estimate in practice. This introduces additional training cost and may degrade the generalizability of the original M/LLMs. \ours aims to estimate uncertainty of a frozen MLLM by employing lightweight post-hoc uncertainty heads, resulting in efficient uncertainty estimation.

\paragraph{Uncertainty decomposition.}
A classical line of work decomposes predictive uncertainty into aleatoric uncertainty, which captures data uncertainty (how data itself is uncertain), and epistemic uncertainty (how the trained model is uncertain) \cite{kendall2017uncertainties,gal2016dropout,malinin2020uncertainty}. Our decomposition is fundamentally different in its target and motivation. Rather than separating data noise from model uncertainty, we study semantic predictive uncertainty in open-ended multimodal settings, where the answer space is not explicitly enumerable and multiple semantically valid answers may exist for the same input and context. In our formulation, the context-specific term measures how broadly the context $t$ defines the plausible answer space, while the multiplicity-specific term measures how many of those context-plausible answers remain compatible with the observed input $x$. Thus, our decomposition is not an aleatoric-epistemic decomposition of model uncertainty, but a semantic decomposition of answer-space uncertainty induced by context and input-answer compatibility. We argue that this distinction is important for MLLMs, because uncertainty can remain high even with a fixed frozen model, not because the model parameters are uncertain, but because the context induces many valid answers or the input supports multiple compatible answers under the context.

Concurrent with our work, \citet{xiao2026vl_calibration} proposed a decoupled multimodal uncertainty estimation framework that combines perturbation-based visual grounding confidence \cite{zhang2024vl_uncertainty} with reasoning-side internal confidence based on token entropy. Unlike our method, their formulation is built on heuristic confidence proxies rather than theoretically motivated quantity. For example, there is no specific reason of why the uncertainty should be decomposed into those two quantities, and why each quantity can be measured by the proposed quantities. Furthermore, their approach focuses on reasoning-oriented MLLMs and requires backbone fine-tuning based on GRPO \cite{shao2024deepseekmath}, whereas our method estimates uncertainty for general-purpose MLLMs using lightweight post-hoc modules on top of a frozen backbone.

\section{More discussions on uncertainty decomposition}

\subsection{Why \renyient?}
\label{sec:appendix_differential_entropy}

\begin{figure}[h]
    \centering
    \includegraphics[width=.85\linewidth]{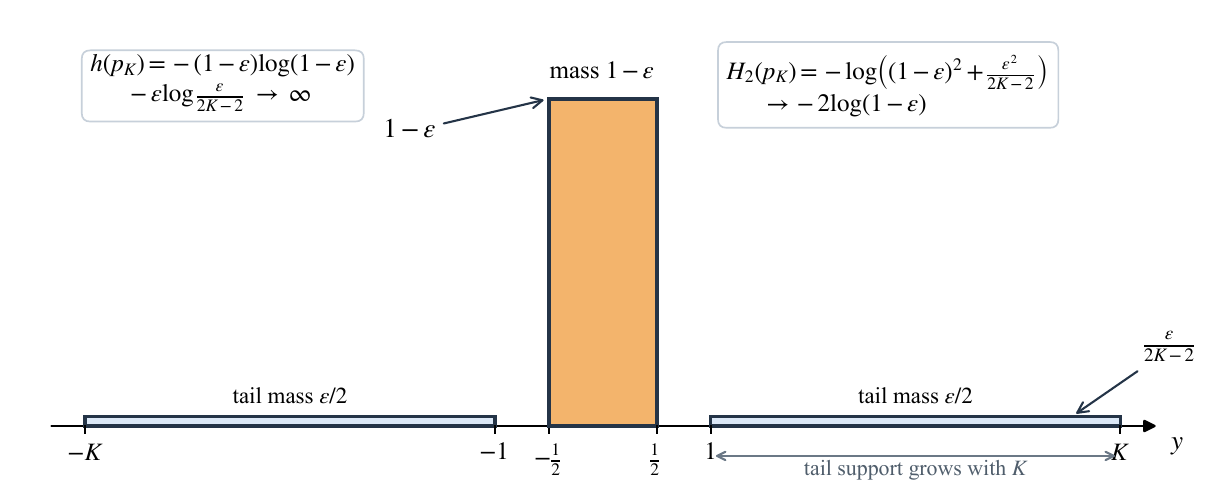}
\caption{\small
{\bf A counterexample for differential entropy.}
The distribution $p_K$ has one dominant region of mass $1-\epsilon$ and diffuse tails with total mass $\epsilon$ spread over growing support. Differential entropy diverges as $K\to\infty$, even though the high-probability semantic region remains unchanged, whereas Rényi-2 remains bounded and converges to $-2\log(1-\epsilon)$.
}    \label{fig:appendix_entropy_counterexample}
\end{figure}

As we discussed in the main paper, when the output space is unknown or continuous, the standard discrete Shannon entropy is not directly applicable.
One may consider the differential entropy ($H_d := \mathbb E_{Y \sim \pi_y} [- \log \pi_Y]$) as an alternative, but it is not aligned with our desired property. More specifically, differential entropy measures the spread of a continuous density with respect to a chosen reference measure. In open-ended answer spaces, any continuous representation of answers is largely arbitrary, so differential entropy would depend on representational choices rather than directly reflecting the multiplicity of semantic answers.

Assume an open-ended answer distribution has one dominant region of plausible answers with a tail of low-probability variants (illustrated in \cref{fig:appendix_entropy_counterexample}). In this case, even though we fix the total probability mass assigned to the tail, by enlarging the tail support, the differential entropy is increased without bound (See \cref{eq:diff_ent_explode}). Instead, a proper uncertainty estimate should more directly reflect the concentration of probability mass on the core plausible answers.
\begin{equation}
\label{eq:diff_ent_explode}
H(p(x)) = -\int p(x) \log p(x) dx = - (1 - \varepsilon) \log (1 - \varepsilon) - \varepsilon \log \frac{\varepsilon}{2k-2} \quad \text{if} \ k\to \infty, H \to \infty
\end{equation}

On the other hand, \renyitwoent remains bounded even when its tail spreads over an increasingly wide support (See \cref{eq:reyni_ent_explode}); whereas differential entropy diverges in the same setting. Therefore, it better reflects the concentration of probability mass on the core plausible answers. 
\begin{equation}
\label{eq:reyni_ent_explode}
H_2(p(x)) = -\log \int p(x)^2 dx = -\log((1 - \varepsilon)^2 - \frac{\varepsilon^2}{2k-2}) \quad \text{if} \ k\to \infty, H_2 \to -2 \log (1-\varepsilon)
\end{equation}

\subsection{The full derivation of uncertainty decomposition with \renyitwoent}
\label{sec:appendix_derivation_renyi}

We denote $s(x, t, y) := p(m=1 \mid x, t, y)$ for simplicity.
We also assume that $p(y \mid t)$ is a uniform distribution over a restricted region $\Omega$, namely $p(y \mid t) = \frac{1}{\int_\Omega dy} = \frac{1}{K_t}$. Now, we define the first and the second moments of the matching probability over $p(y \mid t)$ as follows:
\begin{align}
\begin{split}
\label{eq:energy}
Z_1(x,t) &:= \mathbb E_{Y \sim p(\cdot \mid t)} \big[ s(x, t, Y) \big] = \int s(x,t,y) p(y \mid t) d y\\
Z_2(x,t) &:= \mathbb E_{Y \sim p(\cdot \mid t)} \big[ s(x, t, Y)^2 \big] = \int s(x,t,y)^2 p(y \mid t) d y =\frac{1}{K_t} \int s(x,t,y)^2 d y
\end{split}
\end{align}
From the definition of \renyitwoent, we can derive the following decomposition:
\begin{align}
\begin{split}
\label{eq:uncertainty}
u(x, t) &:= H_2(\pi_y) = - \log \int \pi_y^2 dy = - \log \int \frac{s(x,t,y)^2 p(y \mid t)^2}{Z_1(x,t)^2}dy\\
&= -\log \left[ \frac{1}{K_t^2}\frac{1}{Z_1(x,t)^2} \int s(x, t, y)^2 dy \right]= -\log \left[ \frac{1}{K_t}\frac{Z_2(x,t)}{Z_1(x,t)^2} \right]\\
&= \underbrace{\log K_t}_{:=u_t(t)} + \underbrace{2 \log Z_1(x,t) - \log Z_2 (x,t)}_{:=u_{x \mid t}(x \mid t)}\\
&= \underbrace{\log K_t}_{:=u_t(t)} + \big(\underbrace{-\log\left[\frac{\text{Var}(s)}{(\mathbb E (s))^2} + 1\right]}_{:=u_{x \mid t}(x \mid t)}\big) = \underbrace{\log K_t}_{:=u_t(t)} + \big(\underbrace{-\log\left[\text{CV}(s)^2 + 1\right]}_{:=u_{x \mid t}(x \mid t)}\big),
\end{split}
\end{align}
where $\text{CV}(\cdot)$ denotes the coefficient of variance (also known as normalized root-mean-square deviation), which is defined as follows:
\begin{equation}
\text{CV}(s) = \frac{\sqrt{\text{Var}(s)}}{\mathbb E (s)} = \sqrt{\frac{\mathbb E[s^2]}{\mathbb E[s]^2} - 1} = \sqrt{\frac{Z_2}{Z_1^2} - 1}.
\end{equation}
\cref{eq:uncertainty} implies that the uncertainty of an input $x$ with a given task $t$ is given by the summation of (1) how the output space is complex ($\log K_t$), and (2) the negative coefficient of variance (also known as normalized root-mean-square deviation) of the matching probability $s(x, t, y)$. Namely, it measure how the matching probability between the input $x$ and the answer set $\mathcal Y$ under the context $t$ is distributed. Therefore, $u_{x \mid t}$ measures the degree of many-to-many relationships or multiplicity \cite{chun2025multiplicity} of the matching between $x$ and $y$. For example, if we have more plausible answers compatible to the given $x$ under $t$, the uncertainty will be higher.

\subsection{Derivation of Shannon's entropy in a discrete case}
\label{sec:appendix_shannon_entropy_derivation_taylor}
Now, we assume that the answer space $y$ is discrete and explictly enumeratable. In this case, we can derive the following equation from the definition of discrete entropy:
\begin{align}
\begin{split}
\label{eq:shannon_entropy_derivation}
H(\pi_y) &= -\sum_y \pi_y \log \pi_y = -\sum_y \pi_y (\log s + \underbrace{\log p(y \mid t)}_{\text{constant}} - \log Z_1(x,t))\\
&= \log K_t -\sum_y \pi_y \log s + \sum_y \pi_y \log \underbrace{Z_1(x,t)}_{=\mathbb E [s]} = \log K_t -\sum_y \pi_y \log s + \log( \mathbb E_{Y \sim p(\cdot \mid t)} [s])
\end{split}
\end{align}
Note that $\sum_y \pi_y \log \mathbb E [s] = \log \mathbb E [s] \sum_y \pi_y = \log \mathbb E [s]$, because $\mathbb E [s]$ is no longer a function of $y$.
Here, we rewrite $-\sum_y \pi_y \log s$ as follows:
\begin{align}
-\sum_y \pi_y \log s = -\frac{1}{Z_1}\sum_y s\log s \cdot p(y \mid t) = -\frac{1}{Z_1}\mathbb E_{Y \sim p(\cdot \mid t)} [s \log s].
\end{align}
Therefore, we have $H(\pi_y) = \log K_t + \log Z_1 - \frac{1}{Z_1} \mathbb E_{Y \sim p(\cdot \mid t)}[s \log s]$.
We approximate this value using a second-order Taylor expansion. Let $\mu=\mathbb E [s]$ and $s = \mu + \delta$, where $\mathbb E [\delta] = 0, \mathbb E[\delta^2] = \sigma^2$. Now, apply the Taylor expansion to $f(s) = s \log s$ and take the expectation as follows:
\begin{align}
\begin{split}
\label{eq:shannon_entropy_derivation_taylor}
\mathbb E_{Y \sim p(\cdot \mid t)} &[s\log s] = \mathbb E [f(\mu + \delta)] \approx \mathbb E \left[ \mu \log \mu + (1 + \log \mu) \delta + \frac{\delta^2}{2\mu} \right]\\
&= \mu \log \mu + \mathbb E [\delta] (1 + \log \mu) + \frac{\mathbb E[\delta^2]}{2 \mu} = \mu \log \mu + \frac{\sigma^2}{2\mu} = Z_1 \log Z_1 + \frac{Z_2 - Z_1^2}{2 Z_1}.
\end{split}
\end{align}
Substituting this into \cref{eq:shannon_entropy_derivation}, we have:
\begin{align}
\begin{split}
\label{eq:shannon_entropy_derivation_approx}
H(\pi_y) &= \log K_t + \log Z_1 - \frac{1}{Z_1} \mathbb E_{Y \sim p(\cdot \mid t)}[s \log s]\\
&= \log K_t + \log Z_1 - \log Z_1 - \frac{1}{2}\frac{Z_2}{Z_1^2} = \log K_t - \frac{1}{2}\left(\frac{Z_2}{Z_1^2} - 1\right)
\end{split}
\end{align}
Interestingly, \cref{eq:shannon_entropy_derivation_approx} is very similar to the formulation derived from \renyitwoent in \cref{sec:appendix_derivation_renyi} (the constant term is omitted):
\begin{align}
H(\pi_y) &\approx \log K_t - \frac{1}{2}\frac{Z_2}{Z_1^2} = \underbrace{\log K_t}_{:=u_t'(t)} + \big( \underbrace{-\frac{1}{2}\frac{\text{Var}(s)}{(\mathbb E[s])^2}}_{:= u'_{x \mid t}(x \mid t)} \big)\\
H_2(\pi_y) &= \log K_t - \log \frac{Z_2}{Z_1^2} = \underbrace{\log K_t}_{:=u_t(t)} + \big(\underbrace{-\log\left[\frac{\text{Var}(s)}{(\mathbb E (s))^2} + 1\right]}_{:=u_{x \mid t}(x \mid t)}\big)
\end{align}
Both imply the same intuition: the uncertainty of the posterior $\pi_y$ is decomposed by (1) the task-dependent uncertainty value $\log K_t$, \ie, how the possible outcome space is large and complex, and (2) the uncertainty value by input and task, related to the normalized standard deviation of the matching probability $p(m=1 \mid x, t, y)$ over all the possible outcome $y$ for a given task $t$, $\text{CV}_{Y \sim p(\cdot \mid t)} [s(x,t,y)]$. The Shannon information entropy is directly related to the normalized standard deviation, while Renyi entropy is related to the log of the value.

\section{Method details}

\subsection{Details of MLLM-as-verifier}
\label{sec:appendix_why_no_train_matching_probability}

Our decomposition requires a matching probability function $p(m=1 \mid x, t, y)$. We can employ a parameterized matching probability module, which is a binary classifier that takes a $(x, t, y)$ triplet as input. This strategy is popular in vision-language models (VLMs) training, \eg, image-text matching (ITM) loss \cite{li2021align,li2022blip,li2023blip2} or directly optimizing the loglikelihood of the matching probability between vision-language embeddings \cite{chun2021pcme,chun2024pcmepp,chun2025prolip}. However, we found that training a reliable matching probability module is difficult for two reasons. First, it easily suffers from the context shortcut similar to the text-shortcut problem in early visual-question answering (VQA) tasks \cite{agrawal2018don,cadene2019rubi,clark2019don,bahng2019rebias}. Also, we need a large-scale $(x, t, y)$ triplet dataset to train a matching probability module; however, we empirically found that a matching probability module trained on a triplet dataset, such as visual instruction tuning dataset \cite{llava,tong2024cambrian} does not generalize to arbitrary open-ended settings.

For example, we can achieve this by using a contrastive loss \cite{radford2021clip}, \ie, treating the answers corresponding to the original $x, t$ as positives and all the other answers from the other $x, t$ in the mini-batch as negatives. However, we found that 84\% the questions in the Cambrian dataset are unique, \ie, if we apply contrastive learning to this dataset, the learned module is highly motivated to simply determine the matching based on $t$, while ignoring $x$.

\begin{table}[h]
    \centering
    \small
    \caption{\small {\bf Unique answers per question in Cambrian dataset \cite{tong2024cambrian}.}}
    \label{tab:appendix_cambrian_answers_questions}
\begin{tabular}{@{}lccccccc@{}}
\toprule
\# Unique answers & 1 & 2 & 3--5 & 6--10 & 11--20 & 21--50 & >50 \\ \midrule
\# samples & 811,806 & 120,364 & 22,470 & 7,038 & 2,615 & 945 & 250 \\ \bottomrule
\end{tabular}
\end{table}

As a probing example, we train a simple MLP that estimates whether the given $x, t$ and $y$ are matched upon CLIP-L/14 features \cite{radford2021clip}. When we test the module on the separated validation split with the same distribution with the training dataset, it shows 0.97 AUROC; however, when we mix the questions across the validation dataset, so each test question now can appear with random answers, the AUROC drops significantly to 0.51, which is almost random. In other words, it is very difficult to train matching probability module because the text short between $t$ and $y$ could happen.

\subsection{Prompt for verification-based matching probability}
\label{sec:appendix_matching_verification_template}

\cref{fig:appendix_prompt_matching_prob} shows the prompt template for our MLLM-as-verifier matching probability.

\begin{figure}[h]
\centering
\begin{tcolorbox}[
    colback=gray!3,
    colframe=gray!60,
    title=\textbf{Prompts for matching probability verification},
]
\tt \small
Image: \textcolor{red}{\{x\}}\\
Question: \textcolor{red}{\{t\}}\\
Proposed Answer: \textcolor{red}{\{y\}}\\
Only answer Yes if the image clearly supports the answer.\\ If the image is missing, unclear, or the answer is not visually confirmed, answer No.
\\
Answer Yes or No.
\end{tcolorbox}
\caption{\small {\bf Prompt for MLLM-based matching probability.}}
\label{fig:appendix_prompt_matching_prob}
\end{figure}
Using this template, we measure the $p(y=\text{``Yes''} \mid x, t)$ and $p(y=\text{``No''} \mid x, t)$ to compute the matching probability using \cref{eq:verify_matching_probability}.

\subsection{Matching probability calibration}
\label{sec:appendix_beta_gamma_calibration}

Recall that \cref{eq:verify_matching_probability} needs affine transform variables, $\beta$ and $\gamma$. This affine transform does not change the order of matching probabilities (because this is a monotonic increasing function), but it calibrates the degree of matching probability; smaller $\beta$ makes the distribution of matching probability smoother, and $\gamma$ controls a ``reference point'' for matching probability (\ie, it shifts the distribution).

We apply a simple optimization to get a calibrated affine transform. We first binarize the distribution using a threshold, \ie, $m = 1$ if $p(m) > 0.5$, otherwise $0$, and search for $\beta, \gamma$ that achieves the best matching accuracy in the validation dataset. We use the MMMU validation split for the calibration; because MMMU is a MCQ dataset, there exist positive samples (where $m=1$) and negative samples (where $m=0$) for each $x$ and $t$. We achieve the best matching accuracy with $\beta=0.2$ and $\gamma=-0.1$ for Qwen3VL-2B-Instruct. We use the same value for all the other backbones, such as Qwen3VL-4B and 8B models, and the InternVL model.

\begin{figure}[h]
    \centering
    \begin{subfigure}[b]{0.49\textwidth}
        \centering
        \includegraphics[width=\linewidth]{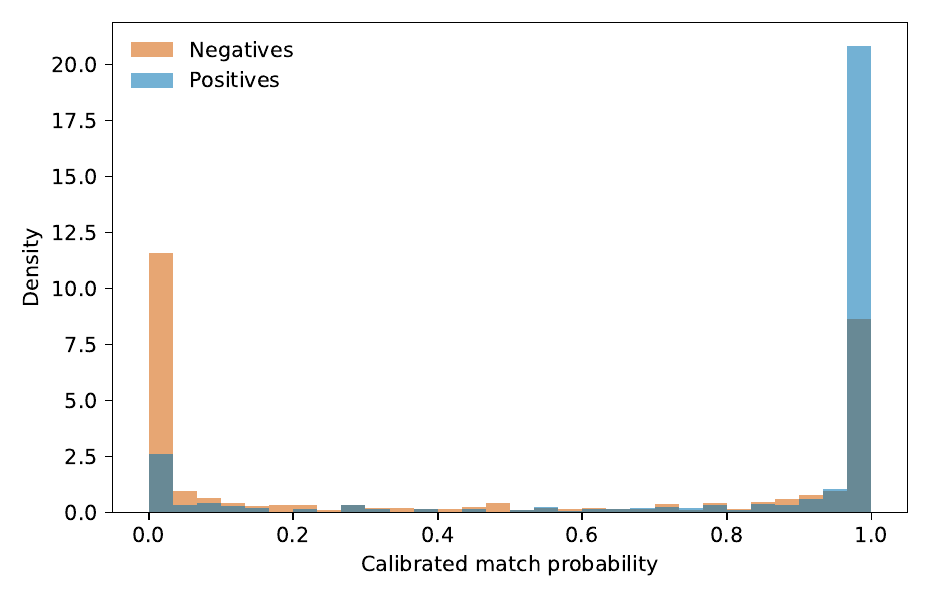}
        \caption{\small Histogram before calibration}
    \end{subfigure}
    \begin{subfigure}[b]{0.49\textwidth}
        \centering
        \includegraphics[width=\linewidth]{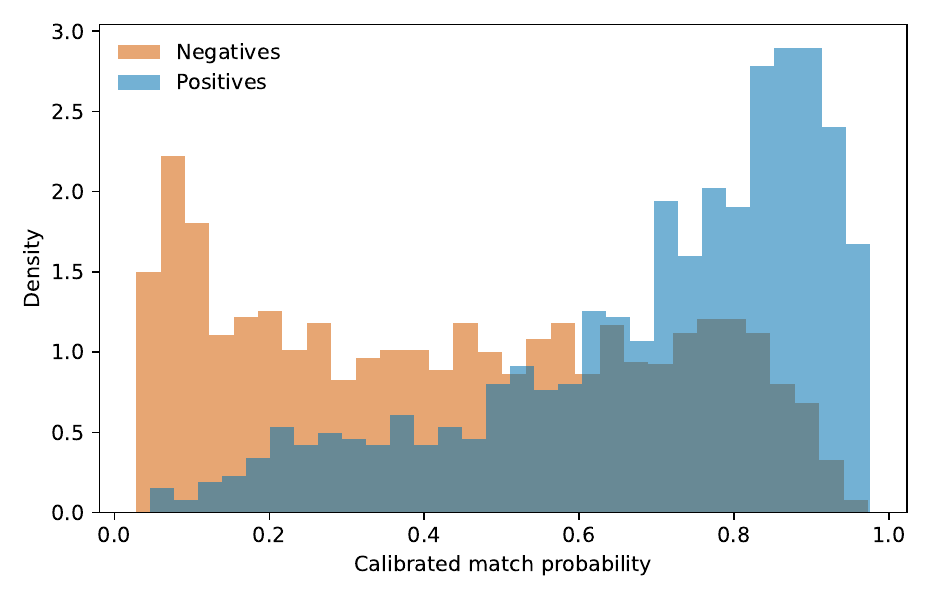}
        \caption{\small Histogram after calibration}
    \end{subfigure}
    \caption{\small {\bf Matching probability histogram before and after calibration.} We visualize the matching probability distributions in the MMMU \cite{mmmu} validation split. The blue bars denote the matching probability distribution for the triplet $(x, t, y+)$ where $y+$ is the corresponding answer among the given multiple-choice questions. The orange bars denote the matching probability when a negative answer is given.
    }
    \label{fig:appendix_matching_probabilty}
\end{figure}

\cref{fig:appendix_matching_probabilty} shows the distribution of matching probabilities on the MMMU validation split before calibration ($\beta=1, \gamma=0$) and after calibration ($\beta=0.2$ and $\gamma=-0.1$). As we discussed previously, the calibration makes the overall matching probability distribution smoother.

In a later section, we will show that this calibration strategy achieves a better uncertainty estimation in the MMMU validation split (where we optimize the hyperparameter) as well as the target benchmark -- See \cref{tab:beta_gamma_ablation} for more details.

\subsection{Dataset collection details}
\label{sec:appendix_dataset_collection}

\begin{figure}
\begin{lstlisting}[language=Python]
def greedy_clustering(dataset, sim_fn, thres):
    """dataset: (x, t, y) triplets
       sim_fn: text similarity function (higher is more similar) """
    # Collect answers by question/context.
    candidates = {}
    for x, t, y in dataset:
        candidates[t].append(y)

    # Process frequent questions first.
    candidates = sorted(candidates.items(), key=lambda item: -len(item[1]))

    clusters, cluster_assignments = [], {}
    for t, answers in candidates:
        # Greedy cluster assignment to the first similar cluster.
        assigned = False
        for cluster in clusters:
            if sim_fn() > thres:
                cluster_assignments[cluster].extend(answers)
                assigned = True
                break

        if not assigned:
            # Ensure that the order of clusters is sorted by the frequency.
            clusters.append(t)
            cluster_assignments[t] = answers
    return cluster_assignments
\end{lstlisting}
\caption{\small {\bf Greedy clustering Python pseudo-code.}}
\label{fig:appendix_algorithm_clustering}
\end{figure}

For dataset collection, we first perform the text clustering described in \cref{fig:appendix_algorithm_clustering} to get plausible answers for each question. We compute the similarity between texts using the CLIP-L/14 text encoder \cite{radford2021clip}. After that, we randomly select 200,000 rows from the Cambrian dataset. For each example $(x_i,t_i,y_i^+)$ assigned to the prototype $p_i$, we construct a candidate set $\mathcal Y_i$ containing the original answer and prototype positives (\ie, answers corresponding to the other questions in the same cluster). Due to the computation complexity, we restrict the size of $\mathcal Y_i$ to 32 in our experiments.

Then, we extract the following features for our training: (1) $f_\theta(x_i, t_i)$: the MLLM feature of the input pair $x_i$ and $t_i$, (2) $f_\theta(t_i)$: the MLLM feature of the input $t_i$, (3) $\log p(\text{``Yes''} \mid x_i, t_i, y_i+)$ for \cref{eq:verify_matching_probability} -- this value is computed for all candidates $y_i \in \mathcal Y_i$; we also extract the log probability of ``No'', and (4) $\log p(y_i \mid t_i)$ for all candidates $y_i \in \mathcal Y_i$. These values are used to estimate the targets $\widehat Z_1(x, t)$, $\widehat Z_2(x, t)$ and $\widehat u(x, t)$, and for input features of the uncertainty estimation modules $F_{Z_1}, F_{Z_2}$ and $F_{u_t}$. In practice, we pre-extracted features and re-use the pre-computed values for training to reduce the computation cost due to the MLLM inference\footnote{Pre-extracted features are released at \url{https://huggingface.co/datasets/SanghyukChun/CoMet-caches}}.

\subsection{Training and architecture details}
\label{sec:appendix_architecture}

We employ three neural uncertainty estimation modules $F_{Z_1}(x, t)$, $F_{\Delta}(x, t)$ and $F_{u_t}(t)$ that estimate $\log \widehat Z_1(x, t)$, $\Delta = \log \widehat Z_1(x, t) - \log \widehat Z_2(x, t)$ and $\widehat u_t(t)$, respectively. Here, $\Delta$ is always nonnegative because $0 \le p(m=1 \mid x, t, y) \le 1$.
We use a lightweight Transformer architecture for the uncertainty modules. We use a Transformer for estimating $F_{Z_1}(x, t)$ and $F_{\Delta}(x, t)$ by taking the MLLM features of $(x, t)$ and $t$ ($f_\theta(x, t)$ and $f_\theta$(t), respectively) for their input. $F_{u_t}(t)$ is also a Transformer, which takes $f_\theta (t)$ as its input. These modules are all 3-layer multi-head Transformers \cite{transformer} with a hidden dimension of 768 and 8 attention heads. The outputs are pooled by an attention pooling, and then fed into a linear head to estimate the target value; $F_{Z_1}(x, t)$ and $F_{\Delta}(x, t)$ share the Transformer backbone, but use different attention pooling modules and the linear projection. Since $Z_1$ and $Z_2$ are the first and second momentum of the probability $\in [0, 1]$, $F_{Z_1} \leq 0$ and $F_{\Delta} \geq 0$;
we use a softplus operation to ensure their value range. This is a similar technique to enforce positive variance values in probabilistic embeddings \cite{chun2021pcme,chun2024pcmepp,chun2025prolip}.

We use a mixture of the pooled features as the input of our module. More specifically, $F_{Z_1}$ takes $f(x, t)$, the MLLM feature of multimodal input $(x, t)$, and $f(t)$, the MLLM feature of text-only input $(t)$. We use four pooled feature types for these features: (1) the last output token -- which is also known as the ``end-of-sentence (EOS)'' token \cite{radford2021clip}, (2) the mean of the last eight output tokens, (3) the mean of the last 32 output tokens, and (4) the mean of all output tokens.
We pre-extract these values from the dataset and cache them to avoid expensive MLLM forward operations.
These four features are fed into the Transformer; $F_{Z_1}$ and $F_{\Delta}$ take 16 token features as their input (four features from each $f(x, t), f(t), f(x, t) - f(t)$, and $f(x, t) \times f(t)$), while $F_{u_t}$ takes four tokens as input.

During training, we optimize the following objective function using a regression loss $\ell$ (\eg, $\ell_1$ loss) on collected data samples described in \cref{sec:p_y_t_dataset_construction}:
\begin{equation}
\label{eq:regression_objective}
\ell( F_{Z_1}(x, t), \log \widehat Z_1(x, t) ) + \ell( F_{\Delta}(x, t), \Delta ) + \ell( \lambda F_{u_t}(t) + F_{Z_1}(x, t) + F_{\Delta}(x, t), \hat u(x, t)).
\end{equation}
Since our training dataset has a different distribution from the actual target benchmarks, we found that the modules trained on our collected training dataset often suffer from the generalizability problem. To mitigate this issue, we additionally employ a lightweight adaptation to a small number of data samples related to the target benchmark. We only updated the last linear layer to estimate each scalar value (hidden dimension $\rightarrow$ 1) and the attention pooling parameters during this adaptation process. Let $\mathcal I^+$ denote the set of samples $(x_i, t_i, y_i)$ that are correctly predicted by an MLLM and $\mathcal I^-$ denote the set of incorrectly predicted samples. The adaptation loss is defined as follows:
\begin{equation}
\label{eq:adaptation_objective}
\mathbb E_{i \in \mathcal I^+, j \in \mathcal I^-} \left[\text{softplus} (u_i' - u_j') \right], \quad \text{where} \ \ u_i' := \lambda F_{u_t}(t_i) + F_{Z_1}(x_i, t_i) +F_{\Delta}(x_i, t_i).
\end{equation}
\cref{eq:adaptation_objective} calibrates the predicted uncertainty value using the actual prediction of an MLLM by enlarging the gap between the uncertainty of correctly predicted samples and that of incorrect samples. For our experiments, we use 900 multiple-choice question VQA samples from the MMMU dataset \cite{mmmu} and 10k open-ended VQA samples for the adaptation from the VQA v2 dataset \cite{vqav2}.

\section{More Experimental Setup}
\label{sec:appendix_more_exp_settings}
\subsection{Evaluation dataset details}
\label{sec:appendix_evaluation_benchmark}

\begin{table}[t]
\centering
\small
\setlength{\tabcolsep}{4pt}
\caption{\small {\bf Overview of evaluation benchmarks.} The detailed statics of the datasets are shown. We report the number of samples for each dataset and its answer type. We also report the performance of Qwen3VL-2B, -4B, -8B-Instruct models and InternVL3.5-1B model on each dataset. For open-ended VQAs, we report the original VQA score and separately report the threshold-based accuracy used in our experiments in the brackets. $\dagger$: The original VQA v2 has 214,354 images, but we randomly select 10k samples for our evaluation. $\ddagger$: These samples are used for the adaptation. $^*$: These samples are used for model selection and hyperparameter tuning.}
\label{tab:appendix_dataset_details}
\begin{tabular}{@{}lllllll@{}}
\toprule
 & \multicolumn{2}{c}{Dataset stats} & \multicolumn{4}{c}{Performance} \\ \midrule
 & \# Samples & Answer type & Qwen 2B & Qwen 4B & Qwen 8B & Intern 1B \\ \midrule
VQA v2 (Validation) \cite{vqav2} & 10,000$^\dagger$ & Open-ended & 62.2 (61.6) & 82.8 (82.5) & 71.6 (71.3) & 76.5 (76.0) \\
VQA v2 (Adaptation) & 10,000$^\ddagger$ & Open-ended & \multicolumn{1}{c}{-} & \multicolumn{1}{c}{-} & \multicolumn{1}{c}{-} & \multicolumn{1}{c}{-} \\
Vizwiz (Validation) \cite{vizwiz} & 4319 & Open-ended & 38.9 (37.2) & 46.4 (45.0) & 40.8 (39.3) & 41.2 (39.0) \\
OK-VQA (Validation) \cite{okvqa} & 5046 & Open-ended & 46.1 (49.3) & 61.2 (64.8) & 55.4 (58.5) & 54.9 (59.0) \\
HallusionBench \cite{hallusionbench} & 1129 & Yes / No & 40.4 & 61.2 & 61.5 & 45.6 \\
MMMU (Test) \cite{mmmu} & 10500 & MCQ & 26 & 43.4 & 47.5 & 37.4 \\
MMMU (Validation) & 900$^\ddagger$$^*$ & MCQ & \multicolumn{1}{c}{-} & \multicolumn{1}{c}{-} & \multicolumn{1}{c}{-} & \multicolumn{1}{c}{-} \\
MMMU (Dev) & 150$^*$ & MCQ & \multicolumn{1}{c}{-} & \multicolumn{1}{c}{-} & \multicolumn{1}{c}{-} & \multicolumn{1}{c}{-} \\
MMMU Pro \cite{mmmu_pro} & 1730 & MCQ & 20.6 & 32.2 & 35.9 & 28 \\
MMStar \cite{mmstar} & 1500 & MCQ & 41.5 & 48.3 & 49.3 & 39.1 \\ \bottomrule
\end{tabular}
\end{table}

We report the details of each dataset in \cref{tab:appendix_dataset_details}. Note that the uncertainty estimation methods compared in this paper keep the backbone MLLM weight frozen; therefore, their performance on each dataset are identical each other. In the main paper, we use 10,000 randomly sampled VQA v2 samples for reporting the VQA v2 results to reduce the expensive inference cost by generation-based methods ($^\dagger$ in the table). We also use randomly selected (but mutually exclusive) 10k samples from VQA v2 and 900 MMMU validation samples for our lightweight adaptation ($^\ddagger$ in the table). Finally, 150 MMMU dev samples and 900 MMMU validation samples are used for model selection and hyperparameter tuning along with the 1k validation split from our collected samples (\cref{sec:p_y_t_dataset_construction}).

In the table, the performances are measured by (1) a soft VQA score \cite{vqav2} for open-ended VQAs and (2) accuracy for the other tasks. Here, we additionally report the threshold-based accuracy for open-ended VQA tasks, which will be described in the following subsection.

\subsection{Evaluation metric details}
\label{sec:appendix_evaluation_metric}

We use the following three metrics as our main metrics: AUROC, AUPRC and FPR95 (False Positive Rate when the True Positive Rate is 95\%), which are standard metrics to measure the performance of uncertainty estimates \cite{hendrycks2016baseline}.

These metrics are defined based on a binary classifier where the input is uncertainty score and the output is whether prediction is correct or incorrect. However, for open-ended VQA tasks, we use a soft score, named VQA score \cite{vqav2} as $\min\left(\frac{m(y)}{3},1\right)$,
where $m(y)$ is the number of human reference answers that match the model prediction $y$ after normalization. The main numbers in \cref{tab:appendix_dataset_details} are the average VQA score over the samples.

To measure the uncertainty estimation performance of each method, we make the soft VQA score to a binary 0, 1 value using a threshold, \ie, if the score is larger than threshold $\tau$, the score becomes 1, otherwise 0. We report the accuracy measured by this in \cref{tab:appendix_dataset_details} (numbers in brackets).

\subsection{Implementation details}
\label{sec:appendix_implementation_details}

We optimize the uncertainty modules using AdamP \cite{heo2021adamp} with a learning rate of $10^{-5}$, a batch size of $64$, weight decay of $0$, and a dropout ratio \cite{srivastava2014dropout} of $0.1$.
We optimize the module for 50 epochs and choose the best model based on the validation AUROC using the 1k held-out split.
For hyperparameters, we search over $\lambda$ in $\{0.25, 0.5, 1.0\}$ based on the MMMU validation split (MMMU val AUROC $-$ MMMU val FPR95).
Based on this, $\lambda=0.25$ was selected as the best hyperparameter except for Qwen3VL-4B, where its best $\lambda$ is $1.0$.
During training, for each iteration, we randomly sample 16 samples from 32 samples of plausible answers. (described in \cref{sec:p_y_t_dataset_construction}).

For the lightweight adaptation stage, we search over batch size in $\{128,256,512\}$ and learning rate in $\{10^{-3},5\times 10^{-4},10^{-4},5\times 10^{-5}\}$ for each trained uncertainty module. Adaptation is also run for up to 50 epochs, and the best checkpoint is selected based on development-set AUROC \footnote{The trained uncertainty modules are released at \url{https://huggingface.co/SanghyukChun/CoMet}}.

\subsection{Comparison methods details}
\label{sec:appendix_comparison_methods}

\begin{figure}[t]
\centering
\begin{tcolorbox}[
    colback=gray!3,
    colframe=gray!60,
    title=\textbf{Prompts for comparison methods},
]
\tt \small
{\bf \textcolor{blue}{[Loglikelihood]}}\\
Image: \textcolor{red}{\{x\}}\\
Question: \textcolor{red}{\{t\}}\\
Proposed Answer: \textcolor{red}{\{y\}}\\
\\
{\bf \textcolor{blue}{[Verbalized Confidence]}}\\
Image: \textcolor{red}{\{x\}}\\
Question: \textcolor{red}{\{t\}}\\
Proposed Answer: \textcolor{red}{\{y\}}\\
How likely is the possible answer to be correct? Briefly explain, then write only one score from 0 to \{confidence\_scale\} inside <confidence></confidence> tags.
\\
\\
{\bf \textcolor{blue}{[P(Correct)]}}\\
Image: \textcolor{red}{\{x\}}\\
Question: \textcolor{red}{\{t\}}\\
Proposed Answer: \textcolor{red}{\{y\}}\\
Is the proposed answer correct?\\
(A) True\\
(B) False\\
The proposed answer is:
\end{tcolorbox}
\vspace{-.25em}
\caption{\small {\bf Prompts for comparison methods.}}
\label{tab:appendix_prompt_comparison_methods}
\vspace{-.75em}
\end{figure}

\textbf{Negative log-likelihood (NLL).} NLL uses the likelihood of the generated answer under the backbone model as a confidence proxy. Lower likelihood corresponds to higher uncertainty. This is a natural decoder-based baseline but is often sensitive to output length and tokenization.
\vspace{-.5em}
\begin{equation}
\label{eq:nll}
\ell(x,t,y) := -\sum_{i=1}^{|y|}\log p_\theta(y_i \mid x,t,y_{<i}).
\end{equation}
In our experiments, however, we found that NLL actually does not serve as a proper uncertainty estimate; it usually shows a very low AUROC score, which means that higher likelihood corresponds to higher uncertainty. This is contradictory to our intuition, and it supports that naively using MLLM output as uncertainty estimate may not work very well in practice.

\textbf{P(Correct).} Following \citet{kadavath2022p_correct}, we ask the model a follow-up verification question about its own answer and use the probability of the affirmative response as a confidence score (See \cref{tab:appendix_prompt_comparison_methods} for the prompt). This baseline is closely related to our verifier, but it is applied to the model's final generated answer rather than to a candidate-answer bank used to construct uncertainty moments.

\textbf{Verbalized confidence.} We also compare against verbalized confidence \cite{verbalized_confidence}, where the model is prompted to explicitly state a confidence value in natural language. This baseline tests whether self-reported confidence alone is sufficient for reliable uncertainty estimation.

\textbf{VL-Uncertainty.} VL-Uncertainty \cite{zhang2024vl_uncertainty} measures uncertainty from the stability of model outputs under multimodal perturbations. It is a strong multimodal baseline, but it is substantially more expensive than our single-pass estimator because it requires multiple perturbed forward passes and repeated generation. 
In our implementation, we use five perturbation rounds, pairing image blur radius $\{0.6,0.8,1.0,1.2,1.4\}$ with question-rephrasing temperatures $\{0.1,0.2,0.3,0.4,0.5\}$. For each perturbed input, we sample an answer with temperature $1.0$, top-$p=0.9$, and at most 32 new tokens, cluster the sampled answers into semantic groups, and use the entropy of the cluster-count distribution as the uncertainty score.

\textbf{Consistency-based baselines.} Following \citet{wang2022self_consistency}, we sample multiple answers for the same input and build an empirical answer distribution from the normalized outputs. We then compute either the entropy of this distribution or its maximum probability. These baselines are effective when short answers repeat reliably, but they become less stable when valid answers can be paraphrased in many ways.
In our experiments, we sample five answers with temperature $1.0$, top-$p=0.9$, and at most 32 new tokens. Then we count the normalized output to make a multinormal distribution; entropy and maximum probability are used for uncertainty score.

\textbf{Candidate-based baselines.} For MCQ benchmarks, we additionally report candidate-based baselines computed over the explicit answer options. More specifically, we compute the log probability of each candidate (\cref{eq:nll}) and use the value as the logit of the candidate; then we get a softmax probability using the logit values. These baselines are not candidate-free, but they provide a useful reference point in settings where the answer set is fully specified at test time.

We consider SteerConf \cite{zhou2025steerconf} as well, which is a prompt-based baseline that queries the model multiple times under different confidence styles (\eg, from very cautious to very confident). SteerConf extracts an answer-confidence pair from each response, and then aggregates these responses into a calibrated uncertainty estimate.
When we applied SteerConf in our settings, it only produces a valid score when the model follows the required output format reliably enough across several steering levels. Concretely, many responses do not contain a parsable answer-confidence pair, therefore, it often fails to predict uncertainty score; we therefore did not compare SteerConf with \ours.

We also did not compare \ours with methods that require additional fine-tuning (or LoRA adaptation \cite{hu2022lora}) on the model, \eg, \citet{kapoor2024large}, ConfTuner \cite{li2025conftuner}, Prompt4Trust \cite{kriz2025prompt4trust}, RLCR \cite{damani2026beyond}, or VL-Calibration \cite{xiao2026vl_calibration}. We aim to keep the original weight of MLLM fixed and measure the uncertainty of the given MLLM without any model steering. These methods may improve calibration by changing the model itself, often with extra supervision, optimization, and hyperparameter tuning, and are therefore not directly comparable to a frozen-weight post-hoc uncertainty estimator.

Finally, it is worth noting that \ours does not need to generate $y$, but only takes $x$ and $t$ as its input for uncertainty estimation (\cref{sec:appendix_architecture}) during inference time, resulting in more efficient uncertainty estimation than the comparison methods.

\section{More Experiment Results}

\subsection{Full results}
\label{sec:appendix_full_results}

\begin{table}[t!]
    \caption{\small {\bf Open-ended VQA results.} We report AUROC (AUC), AUPRC (AP), and FPR@95 of uncertainty estimation methods on open-ended VQA benchmarks. ``GF'' denotes generation-free.}
    \label{tab:openset_vqa}
    \centering
    \footnotesize
\setlength{\tabcolsep}{3pt}
\begin{tabular}{@{}llllllllllllll@{}}
\toprule
 &  & \multicolumn{3}{c}{Qwen3VL-2B-Instruct} & \multicolumn{3}{c}{Qwen3VL-4B-Instruct} & \multicolumn{3}{c}{Qwen3VL-8B-Instruct} & \multicolumn{3}{c}{InternVL3.5-1B} \\ \midrule
Method & GF & AUC $\uparrow$ & AP $\uparrow$ & FPR $\downarrow$ & AUC $\uparrow$ & AP $\uparrow$ & FPR $\downarrow$ & AUC $\uparrow$ & AP $\uparrow$ & FPR $\downarrow$ & AUC $\uparrow$ & AP $\uparrow$ & FPR $\downarrow$ \\
\midrule
\multicolumn{14}{c}{VQA v2 (10k) \cite{vqav2}} \\ \midrule
Consist. (Ent) & \nomark & .624 & .471 & .925 & .677 & .266 & .896 & .684 & .423 & .891 & .535 & .249 & .956 \\
Consist. (Max) & \nomark & .620 & .470 & .914 & .678 & .266 & .896 & .680 & .421 & .891 & .535 & .249 & .956 \\
NLL & \nomark & .531 & .389 & .991 & .425 & .145 & .995 & .450 & .253 & .991 & .547 & .211 & .986 \\
Verbalized Conf & \nomark & .545 & .419 & .913 & .608 & .218 & .877 & .651 & .460 & .808 & .623 & .350 & .846 \\
P(Correct) & \nomark & .600 & .454 & .913 & .724 & .361 & .795 & .740 & .554 & .743 & .610 & .274 & .982 \\
VL Uncertainty & \nomark & .630 & .520 & .849 & .684 & .355 & .776 & .665 & .448 & .836 & .623 & .339 & .878 \\
\rowcolor{gray!20}
\ours (Ours) & \yesmark & \textbf{.833} & \textbf{.765} & \textbf{.609} & \textbf{.844} & \textbf{.524} & \textbf{.644} & \textbf{.836} & \textbf{.664} & \textbf{.664} & \textbf{.773} & \textbf{.542} & \textbf{.709} \\ \midrule
\multicolumn{14}{c}{VizWiz (val) \cite{vizwiz}} \\ \midrule
Consist. (Ent) & \nomark & .594 & .697 & .908 & .640 & .656 & .906 & .634 & .701 & .904 & .475 & .590 & .959 \\
Consist. (Max) & \nomark & .596 & .697 & .908 & .639 & .656 & .906 & .634 & .701 & .904 & .475 & .590 & .959 \\
NLL & \nomark & .567 & .669 & .942 & .497 & .543 & .975 & .427 & .547 & .980 & .566 & .575 & .975 \\
Verbalized Conf & \nomark & .606 & .723 & .823 & .531 & .564 & .895 & .656 & .773 & .727 & .655 & .751 & .786 \\
P(Correct) & \nomark & .756 & .827 & .786 & .718 & .734 & .839 & .715 & .796 & .734 & .734 & .772 & .839 \\
VL Uncertainty & \nomark & .662 & .767 & .837 & .641 & .687 & .842 & .658 & .761 & .806 & .679 & .746 & .854 \\
\rowcolor{gray!20}
\ours (Ours) & \yesmark & \textbf{.783} & \textbf{.849} & \textbf{.732} & \textbf{.793} & \textbf{.813} & \textbf{.696} & \textbf{.815} & \textbf{.865} & \textbf{.663} & \textbf{.752} & \textbf{.813} & \textbf{.759} \\ \midrule
\multicolumn{14}{c}{OK-VQA \cite{okvqa}} \\ \midrule
Consist. (Ent) & \nomark & .655 & .611 & .921 & .666 & .471 & .895 & .653 & .520 & .918 & .477 & .393 & .963 \\
Consist. (Max) & \nomark & .654 & .610 & .921 & .663 & .470 & .895 & .652 & .519 & .918 & .477 & .393 & .963 \\
NLL & \nomark & .609 & .577 & .955 & .553 & .384 & .953 & .548 & .452 & .938 & .615 & .512 & .887 \\
Verbalized Conf & \nomark & .621 & .635 & .790 & .548 & .352 & .916 & .619 & .561 & .826 & .641 & .535 & .888 \\
P(Correct) & \nomark & .698 & .663 & .892 & .693 & .514 & .876 & .689 & .606 & .821 & .658 & .551 & .870 \\
VL Uncertainty & \nomark & .660 & .651 & .869 & .667 & .524 & .817 & .650 & .568 & .853 & .608 & .519 & .876 \\
\rowcolor{gray!20}
\ours (Ours) & \yesmark & \textbf{.740} & \textbf{.738} & \textbf{.781} & \textbf{.715} & \textbf{.557} & \textbf{.813} & \textbf{.746} & \textbf{.656} & \textbf{.776} & \textbf{.677} & \textbf{.586} & \textbf{.844} \\ \bottomrule
\end{tabular}
\end{table}

\begin{table}[t!]
    \centering
    \caption{\small {\bf Unanswerability and hallucination detection results.} Details are the same as \cref{tab:openset_vqa}.}
    \label{tab:unanswer_hallu_detection_results}
    \footnotesize
\setlength{\tabcolsep}{3pt}
\begin{tabular}{@{}llllllllllllll@{}}
\toprule
 &  & \multicolumn{3}{c}{Qwen3VL-2B-Instruct} & \multicolumn{3}{c}{Qwen3VL-4B-Instruct} & \multicolumn{3}{c}{Qwen3VL-8B-Instruct} & \multicolumn{3}{c}{InternVL3.5-1B} \\ \midrule
Method & GF & AUC $\uparrow$ & AP $\uparrow$ & FPR $\downarrow$ & AUC $\uparrow$ & AP $\uparrow$ & FPR $\downarrow$ & AUC $\uparrow$ & AP $\uparrow$ & FPR $\downarrow$ & AUC $\uparrow$ & AP $\uparrow$ & FPR $\downarrow$ \\ \midrule
\multicolumn{14}{c}{VizWiz (unanswerable) \cite{vizwiz}} \\ \midrule
Consist. (Ent) & \nomark & .608 & .408 & .852 & .628 & .437 & .823 & .656 & .443 & .852 & .465 & .307 & .933 \\
Consist. (Max) & \nomark & .607 & .409 & .864 & .628 & .436 & .852 & .654 & .443 & .852 & .465 & .307 & .933 \\
NLL & \nomark & .570 & .383 & .942 & .526 & .329 & .944 & .413 & .272 & .982 & .560 & .341 & .935 \\
Verbalized Conf & \nomark & .615 & .470 & .961 & .553 & .360 & .939 & .621 & .521 & .946 & .727 & .575 & .952 \\
P(Correct) & \nomark & .751 & .576 & .712 & .660 & .421 & .764 & .600 & .397 & .839 & .788 & .594 & \textbf{.628} \\
VL Uncertainty & \nomark & .651 & .450 & .910 & .545 & .367 & .957 & .543 & .359 & .950 & .693 & .481 & .860 \\
\rowcolor{gray!20}
\ours (Ours) & \yesmark & \textbf{.788} & \textbf{.596} & \textbf{.641} & \textbf{.767} & \textbf{.556} & \textbf{.641} & \textbf{.797} & \textbf{.635} & \textbf{.689} & \textbf{.808} & \textbf{.641} & \textbf{.628} \\ \midrule
\multicolumn{14}{c}{HallusionBench \cite{hallusionbench}} \\ \midrule
Consist. (Ent) & \nomark & .564 & .631 & .920 & .569 & .439 & .932 & .572 & .442 & .920 & .522 & .575 & .919 \\
Consist. (Max) & \nomark & .562 & .630 & .920 & .567 & .431 & .957 & .566 & .438 & .920 & .521 & .574 & .919 \\
NLL & \nomark & .443 & .548 & .981 & .567 & .425 & .943 & .585 & .434 & .943 & .468 & .524 & .966 \\
Verbalized Conf & \nomark & .583 & .664 & .945 & .543 & .435 & .921 & .604 & .486 & .876 & .503 & .545 & .956 \\
P(Correct) & \nomark & .594 & .661 & .921 & .636 & \textbf{.527} & \textbf{.836} & \textbf{.670} & \textbf{.536} & .848 & .510 & .512 & .995 \\
VL Uncertainty & \nomark & .417 & .717 & 1.000 & .500 & .393 & 1.000 & .500 & .393 & 1.000 & .429 & .571 & .909 \\
\rowcolor{gray!20}
\ours (Ours) & \yesmark & \textbf{.715} & \textbf{.791} & \textbf{.779} & \textbf{.644} & \underline{.523} & \underline{.881} & \underline{.605} & \underline{.510} & \textbf{.841} & \textbf{.656} & \textbf{.658} & \textbf{.906} \\ \bottomrule
\end{tabular}
\end{table}

\begin{table}[t!]
    \centering
    \footnotesize
    \caption{\small {\bf Multiple Choice Question (MCQ) VQA results.} Details are the same as \cref{tab:openset_vqa}. Note that Candidate (Ent) and (Max) use the candidate set information, while the other methods do not.}
    \label{tab:mcq_vqa}
\setlength{\tabcolsep}{3pt}
\begin{tabular}{@{}llllllllllllll@{}}
\toprule
 &  & \multicolumn{3}{l}{Qwen3VL-2B-Instruct} & \multicolumn{3}{l}{Qwen3VL-4B-Instruct} & \multicolumn{3}{l}{Qwen3VL-8B-Instruct} & \multicolumn{3}{l}{InternVL3.5-1B} \\ \midrule
 &  & AUC $\uparrow$ & AP $\uparrow$ & FPR $\downarrow$ & AUC $\uparrow$ & AP $\uparrow$ & FPR $\downarrow$ & AUC $\uparrow$ & AP $\uparrow$ & FPR $\downarrow$ & AUC $\uparrow$ & AP $\uparrow$ & FPR $\downarrow$ \\ \midrule
\multicolumn{14}{c}{MMMU Test\cite{mmmu}} \\ \midrule
Candidate (Ent) & \yesmark & .732 & .859 & .812 & .700 & .692 & .880 & .692 & .661 & .856 & .559 & .648 & .920 \\
Candidate (Max) & \yesmark & .717 & .850 & .832 & .694 & .680 & .896 & .685 & .647 & .872 & .552 & .642 & .933 \\
Consist. (Ent) & \nomark & .804 & .926 & .511 & .707 & .794 & \textbf{.669} & .695 & .740 & \textbf{.741} & \textbf{.638} & \textbf{.751} & \textbf{.845} \\
Consist. (Max) & \nomark & .796 & .922 & .557 & .705 & .790 & .690 & .691 & .732 & .744 & .634 & .746 & .855 \\
NLL & \nomark & .155 & .575 & .997 & .271 & .431 & .995 & .337 & .432 & .977 & .386 & .539 & .990 \\
Verbalized Conf & \nomark & .666 & .832 & .934 & .521 & .621 & .950 & .552 & .586 & .923 & .521 & .654 & .940 \\
P(Correct) & \nomark & .430 & .680 & .981 & .743 & .775 & .769 & .702 & .696 & .830 & .580 & .691 & .903 \\
VL Uncertainty & \nomark & .837 & .941 & .404 & .696 & .770 & .727 & .697 & .730 & .762 & .579 & .692 & .915 \\
\rowcolor{gray!20}
\ours (Ours) & \yesmark & \textbf{.875} & \textbf{.956} & \textbf{.350} & \textbf{.777} & \textbf{.819} & \underline{.688} & \textbf{.770} & \textbf{.783} & \textbf{.725} & \underline{.630} & \underline{.730} & \underline{.880} \\ \midrule
\multicolumn{14}{c}{MMMU Pro \cite{mmmu_pro}} \\ \midrule
Candidate (Ent) & \yesmark & .725 & .903 & .775 & .672 & .803 & .853 & .680 & .784 & .807 & .598 & .770 & .934 \\
Candidate (Max) & \yesmark & .715 & .898 & .800 & .659 & .786 & .887 & .672 & .770 & .850 & .583 & .763 & .929 \\
Consist. (Ent) & \nomark & .769 & .928 & .612 & .696 & .847 & \textbf{.670} & .708 & \textbf{.831} & \textbf{.695} & \textbf{.662} & \textbf{.814} & \textbf{.884} \\
Consist. (Max) & \nomark & .757 & .921 & .635 & .694 & .845 & .700 & .705 & .826 & .700 & .648 & .801 & .916 \\
NLL & \nomark & .192 & .659 & .996 & .293 & .554 & .995 & .359 & .569 & .968 & .439 & .673 & .993 \\
Verbalized Conf & \nomark & .626 & .834 & .946 & .493 & .713 & .961 & .564 & .677 & .925 & .521 & .741 & .928 \\
P(Correct) & \nomark & .522 & .781 & .970 & .711 & .828 & .758 & .704 & .802 & .760 & .555 & .747 & .935 \\
VL Uncertainty & \nomark & .815 & .949 & .448 & .703 & .839 & .730 & .714 & .825 & .709 & .652 & .810 & .873 \\
\rowcolor{gray!20}
\ours (Ours) & \yesmark & \textbf{.834} & \textbf{.953} & \textbf{.442} & \textbf{.739} & \textbf{.851} & {.778} & \textbf{.726} & {.811} & {.794} & {.610} & {.771} & {.931} \\ \midrule
\multicolumn{14}{c}{MMStar \cite{mmstar}} \\ \midrule
Candidate (Ent) & \yesmark & .701 & .643 & .810 & .669 & .533 & .849 & .654 & .492 & .874 & .590 & .554 & .909 \\
Candidate (Max) & \yesmark & .692 & .632 & .803 & .667 & .524 & .858 & .651 & .489 & .879 & .592 & .548 & .921 \\
Consist. (Ent) & \nomark & .753 & .844 & .564 & .723 & .797 & .574 & .740 & .800 & .539 & \textbf{.741} & \textbf{.848} & \textbf{.574} \\
Consist. (Max) & \nomark & .746 & .837 & .567 & .719 & .786 & .644 & .732 & .785 & .622 & .739 & .847 & \textbf{.574} \\
NLL & \nomark & .226 & .434 & .995 & .177 & .352 & .999 & .206 & .356 & .996 & .247 & .462 & .999 \\
Verbalized Conf & \nomark & .559 & .651 & .919 & .607 & .694 & .821 & .564 & .658 & .850 & .551 & .691 & .808 \\
P(Correct) & \nomark & .581 & .604 & .958 & .732 & .725 & .763 & .793 & .815 & .581 & .634 & .710 & .896 \\
VL Uncertainty & \nomark & .754 & .838 & .593 & .765 & .827 & .524 & .754 & .811 & .590 & .529 & .644 & .917 \\
\rowcolor{gray!20}
\ours (Ours) & \yesmark & \textbf{.819} & \textbf{.878} & \textbf{.523} & \textbf{.830} & \textbf{.859} & \textbf{.501} & \textbf{.853} & \textbf{.881} & \textbf{.444} & \underline{.739} & \underline{.815} & \underline{.743} \\ \bottomrule
\end{tabular}
\end{table}

We show the full results of open-ended multimodal understanding benchmarks (\cref{tab:openset_vqa}), unanswerability/hallucination detection benchmarks (\cref{tab:unanswer_hallu_detection_results}), and multiple choice question (MCQ) visual question answering (VQA) benchmarks (\cref{tab:mcq_vqa}).

\subsection{Decomposed pre-training targets vs. unified target}
\label{sec:appendix_additional_loss_ablation}

We compare \ours with unified entropy targets without decomposition, but using the same candidates and verifier outputs.
We construct a candidate distribution by applying a softmax to the MLLM-as-verifier results, $s_y = \log \frac{p(\text{Yes} | x, t, y)}{p(\text{No} | x, t, y)}$ over answer candidates for each question. Then, we train the same uncertainty module to regress either Shannon entropy $-\sum_y p_y \log p_y$ or \renyitwoent $-\log\sum_y p_y^2$ as a single target, where $p_y=\mathrm{softmax}(s_y)$. \cref{tab:entropy_loss_ablation} shows that CoMet achieves 0.798 average AUROC, outperforming Shannon (0.729) and \renyitwo (0.735). These results support the benefit of component-wise pre-training over directly regressing a single entropy target.

\begin{table}[t]
\small
\centering
\caption{\small \textbf{Unified entropy target vs. decomposed target for pre-training.}
Using the same answer candidates and MLLM-as-verifier results, we compare the decomposed pre-training target by \ours with the unified target by Shannon's entropy and \renyitwo entropy.
We use the Qwen3VL-2B backbone for all experiments and use the same dataset, architecture, and adaptation. The other details are the same as \cref{tab:verifier_ablation}.
}
\label{tab:entropy_loss_ablation}
\resizebox{\linewidth}{!}{
\begin{tabular}{lccccccccc}
\toprule
Object & VQA v2 & VW & OK-VQA & VW (U) & HB & MMMU & MMMU Pro & MMStar & Avg \\ \midrule
Shannon Ent. & .747 & .685 & .651 & .721 & .626 & .806 & .779 & .820 & .729 \\
\renyitwo Ent. & .748 & .747 & .677 & .784 & .538 & .821 & .774 & .789 & .735 \\
\rowcolor{gray!20}
\ours & \textbf{.833} & \textbf{.783} & \textbf{.740} & \textbf{.788} & \textbf{.715} & \textbf{.875} & \textbf{.834} & \textbf{.819} & \textbf{.798} \\ \bottomrule
\end{tabular}
}
\end{table}

\subsection{More discussion on MLLM-as-verifier and candidate construction}
\label{sec:appendix_additional_analysis}

\textbf{More discussion on MLLM-as-verifier.}
As we discussed in the main paper, the gap due to the choice of verifiers is not as significant as their disagreements. For example, InternVL3.5-1B and Qwen3VL-8B disagree with Qwen3VL-2B on 35.0\% and 19.3\% of collected candidates, but the average uncertainty estimation AUROC does not drop significantly.
One possible explanation is that our uncertainty targets aggregate matching probabilities over $p(y \mid t)$, rather than directly using individual verification results.
When we aggregate verification results for each question, the disagreement rates with Qwen3VL-2B become 28.0\% for InternVL3.5-1B and 22.6\% for Qwen3VL-8B, narrowing the gap between the two verifiers.
These results support the limited sensitivity to the choice of the verifier model.

\textbf{More discussion on LLM-generated candidate construction.}
As we mentioned in the main paper, LLM-based answer candidate generation is not straightforward for vision-related questions. When an LLM is directly given a question such as ``What is in this image?'', it often refuses to answer, \eg, ``I currently cannot view or analyze images,'' rather than producing plausible visual answers.
Therefore, we need to design a dedicated candidate-generation prompt. \cref{fig:candidate_construction_appendix} shows our dedicated prompt with in-context samples. We use the candidates constructed by our dataset-based method as the in-context examples. Using this setup, we generate 28 candidates for each question using Qwen3-8B.

\begin{figure}[t]
\centering
\begin{tcolorbox}[
    colback=gray!3,
    colframe=gray!60,
    title=\textbf{Prompts for candidate construction study},
]
\tt \small

You are constructing plausible candidate answers for a visual question. Generate exactly one short answer phrase. Do not explain. Do not output multiple answers. Question: \textcolor{red}{\{t\}}, Known valid answers (examples, not exhaustive): \textcolor{red}{\{in-context examples\}}. Generate one additional plausible answer, preferably different from the examples above. Answer:
\end{tcolorbox}
\caption{\small {\bf Prompts for candidate construction study.}}
\label{fig:candidate_construction_appendix}
\end{figure}

\begin{table}[t]
\small
\centering
\caption{\small \textbf{Impact of the number of candidates.}
\ours results by varying the number of pre-training answer candidates are shown. Each question has 32 candidates.
Qwen3VL-2B-Instruct is used for the backbone.
}
\label{tab:n_cadidates_ablation}
\begin{tabular}{lccccccccc}
\toprule
\# questions & VQA v2 & VW & OK-VQA & VW (U) & HB & MMMU & MMMU Pro & MMStar & Avg \\ \midrule
3k & .835 & .741 & .727 & .751 & .707 & .879 & .831 & .813 & .786 \\
5k & .828 & .800 & .738 & .795 & .680 & .874 & .811 & .804 & .791 \\
10k & .828 & .774 & .731 & .797 & .715 & .863 & .794 & .782 & .785 \\
20k & .831 & .775 & .734 & .793 & .700 & .873 & .806 & .808 & .790 \\
25k & .831 & .774 & .738 & .788 & .702 & .874 & .811 & .809 & .791 \\
100k & .823 & .741 & .721 & .711 & .697 & .872 & .816 & .819 & .775 \\
\rowcolor{gray!20}
200k & .833 & .783 & .740 & .788 & .715 & .875 & .834 & .819 & .798 \\ \bottomrule
\end{tabular}
\end{table}

\begin{table}[t!]
\small
\centering
\caption{\small \textbf{Impact of the number of candidates per question.}
$K$ denotes the number of candidate answers for each question. The details are the same as \cref{tab:n_cadidates_ablation}.
}
\label{tab:n_cadidates_ablation2}
\begin{tabular}{lccccccccc}
\toprule
$K$ & VQA v2 & VW & OK-VQA & VW (U) & HB & MMMU & MMMU Pro & MMStar & Avg \\ \midrule
4 & .832 & .779 & .722 & .776 & .681 & .877 & .818 & .795 & .785 \\
8 & .812 & .772 & .709 & .791 & .615 & .866 & .822 & .809 & .775 \\
16 & .835 & .778 & .734 & .777 & .681 & .877 & .833 & .816 & .791 \\
\rowcolor{gray!20}
32 & .833 & .783 & .740 & .788 & .715 & .875 & .834 & .819 & .798 \\ \bottomrule
\end{tabular}
\end{table}

\textbf{The impact of the candidate size.}
\cref{tab:n_cadidates_ablation} and \cref{tab:n_cadidates_ablation2} show the impact of the size of answer candidates. 
\cref{tab:n_cadidates_ablation} shows that increasing the number of questions from 3k to 200k improves average AUROC from 0.786 to 0.798. When the number of candidates per question changed from 4 to 32, the average performance tends to improve (0.785 to 0.798) as shown in \cref{tab:n_cadidates_ablation2}.

\subsection{Impact of adaptation}
\label{sec:appendix_adaptation_ablation}

\cref{tab:appendix_adaptation_ablation} shows the impact of our lightweight adaptation method. When we use the Qwen3VL-8B-Instruct backbone, even without adaptation, \ours already provides a meaningful uncertainty signal compared to strong baselines, such as Verbalized confidence and P(Correct) (average AUROC 0.609 vs. 0.604 and 0.598, respectively). By adapting \ours on the MMMU validation split, \ours ties with VL uncertainty, the strongest baseline in average AUROC. By adding VQA samples for the adaptation, the uncertainty estimation by \ours is significantly improved.

Importantly, the effect depends on the adaptation source. MMMU-only adaptation specializes the uncertainty head to MCQ evaluation (MMMU, MMMU-Pro, and MMStar), while open-ended VQA performances are somewhat worse than the original (\eg, OK-VQA AUROC 0.621 $\rightarrow$ 0.593 after the MMMU-only adaptation). In contrast, VQA-only adaptation significantly improves open-ended VQA tasks, but its MCQ performance is somewhat worse than that of the MMMU-only adaptation. Combining VQA and MMMU adaptation gives the best average performance, suggesting that the two sources provide complementary calibration signals.

In the main paper, we show that using the same samples for pre-training and adaptation also works. More specifically, we evaluate the model's answers on Cambrian using the same VQA scoring protocol and convert the scores to binary correctness labels for adaptation.

\subsection{Parameter study}
\label{sec:appendix_parameter_study}

\cref{tab:lambda_ablation} and \cref{tab:beta_gamma_ablation} show the parameter study results on $\lambda$ and $(\beta, \gamma)$ using the Qwen3VL-8B-Instruct model. We select the best hyperparameter based on the MMMU validation split.
In the tables, we observe that our hyperparameter choice performs the best on the MMMU validation split as well as the average score.

\begin{table}[t]
\centering
\footnotesize
\setlength{\tabcolsep}{4pt}
\caption{\small {\bf Impact of the adaptation strategy.} Measured by the Qwen3VL-8B-Instruct backbone.}
\label{tab:appendix_adaptation_ablation}
\begin{tabular}{@{}lllllllll@{}}
\toprule
Method & VQA v2 & VizWiz & OK-VQA & MMMU Test & MMMU Pro & MMStar & Avg \\ \midrule
\ours (No Adapt) & .503 & .676 & .621 & .671 & .659 & .526 & .609 \\
\ours (MMMU only adapt) & .654 & .730 & .593 & .772 & .731 & .863 & .724 \\
\ours (VQA only adapt) & .826 & .816 & .726 & .702 & .678 & .804 & .759 \\
\rowcolor{gray!20}
\ours (VQA + MMMU adapt) & .836 & .815 & .746 & .770 & .726 & .853 & .791 \\ \midrule
Consist (ent) & .624 & .594 & .655 & .732 & .725 & .701 & .672 \\
Verbalized Confidence & .545 & .606 & .621 & .666 & .626 & .559 & .604 \\
P(Correct) & .600 & .756 & .698 & .430 & .522 & .581 & .598 \\
VL uncertainty & .630 & .662 & .660 & .837 & .815 & .754 & .726 \\ \bottomrule
\end{tabular}
\vspace{-.75em}
\end{table}

\begin{table}[t]
\small
\centering
\caption{\small {\bf Impact of $\lambda$.} We use Qwen3VL-8B-Instruct as the backbone. Hyperparameters are selected on the MMMU validation split (the first column).}
\label{tab:lambda_ablation}
\setlength{\tabcolsep}{4pt}
\begin{tabular}{@{}lcccccccc@{}}
\toprule
$\lambda$ & MMMU val & VQA v2 & VizWiz (val) & OK-VQA & MMMU Test & MMMU Pro & MMStar & Average \\ \midrule
0 & .837 & .834 & .734 & .735 & .753 & .707 & \textbf{.857} & .780 \\
\rowcolor{gray!20}
0.25 & \textbf{.842} & .836 & .815 & .746 & .770 & \textbf{.726} & .853 & \textbf{.798} \\
0.5 & .822 & \textbf{.841} & .796 & .741 & \textbf{.772} & .715 & .835 & .789 \\
1 & .828 & .839 & \textbf{.821} & \textbf{.751} & .767 & .707 & .852 & .795 \\ \bottomrule
\end{tabular}
\vspace{-.75em}
\end{table}

\begin{table}[t!]
\small
\centering
\setlength{\tabcolsep}{4pt}
\caption{\small {\bf Impact of $\beta, \gamma$.} Details are the same as \cref{tab:lambda_ablation}.}
\label{tab:beta_gamma_ablation}
\begin{tabular}{@{}lcccccccc@{}}
\toprule
$\beta/\gamma$ & MMMU val & VQA v2 & VizWiz (val) & OK-VQA & MMMU Test & MMMU Pro & MMStar & Average \\ \midrule
\rowcolor{gray!20}
0.2/-0.1 & \textbf{.842} & .836 & \textbf{.815} & \textbf{.746} & \textbf{.770} & .726 & \textbf{.853} & \textbf{.798} \\
1/0 & .835 & \textbf{.837} & .796 & \textbf{.746} & .760 & .736 & .832 & .792 \\
\bottomrule
\end{tabular}
\vspace{-.25em}
\end{table}

\begin{table}[t]
\small
\setlength{\tabcolsep}{4pt}
\caption{\small \textbf{Intervention on $x$ and $t$.} We intervene on VQA v2 samples to control $u_t$ and $u_{x\mid t}$. (a) used the full 10k evaluation split of VQA v2. For (b) and (c), we use 525 ``how many questions'' and 1,809 ``how many'' and ``what color'' questions. Qwen3VL-2B is used for the backbone.}
\label{tab:intervention}
\begin{subtable}[h]{0.48\textwidth}
\caption{\small \textbf{Intervetion on $x$.} We intervene on the visual input $x$ by adding stronger Gaussian blur.}
\label{tab:intervention_a}
\resizebox{\linewidth}{!}{
\begin{tabular}{lccccc}
\toprule
Radius & 0 (clean) & 2.0 & 4.0 & 8.0 & 16.0 \\ \midrule
$u_t$ & 0.79 & 0.79 & 0.79 & 0.79 & 0.79 \\
$u_{x\mid t}$ & $-2.39$ & $-1.69$ & $-1.42$ & $-1.27$ & $-0.87$ \\
$u$ & $-2.19$ & $-1.50$ & $-1.22$ & $-1.07$ & $-0.68$ \\  \bottomrule
\end{tabular}
}
\end{subtable}
\hfill
\begin{subtable}[h]{0.245\textwidth}
\caption{\small \textbf{Intervention on $t$ with an extreme case.}}
\label{tab:intervention_b}
\resizebox{\linewidth}{!}{
\begin{tabular}{llll}
\toprule
$K$ & $u_t$ & $u_{x \mid t}$ & $u$ \\ \midrule
20 & $1.99$ & $-1.81$ & $-1.31$ \\
30 & $2.04$ & $-1.90$ & $-1.39$ \\
40 & $2.09$ & $-2.25$ & $-1.73$ \\ \bottomrule
\end{tabular}
}
\end{subtable}
\hfill
\begin{subtable}[h]{0.249\textwidth}
\caption{\small \textbf{More realistic $t$ control.}}
\label{tab:intervention_c}
\resizebox{\linewidth}{!}{
\begin{tabular}{lccc}
\toprule
$K$ & $u_t$ & $u_{x \mid t}$ & $u$ \\ \midrule
2 & $-0.55$ & $-2.99$ & $-3.13$ \\
4 & $-0.39$ & $-2.40$ & $-2.50$ \\
8 & $0.02$ & $-1.83$ & $-1.82$ \\
- & $1.79$ & $-2.47$ & $-2.02$ \\ \bottomrule
\end{tabular}
}
\end{subtable}
\end{table}

\begin{table}[t!]
\small
\centering
\setlength{\tabcolsep}{4pt}
\caption{\small \textbf{Additional parameters and memory by \ours.} Bfloat16 is used, excluding activations.}
\label{tab:additional_params_memory}
\begin{tabular}{cccc}
\toprule
Backbone & Backbone param/memory (B/GiB) & \ours param/memory (M/MiB) & Increase (\%) \\ \midrule
Qwen3VL-2B & 2.13 / 3.96 & 50.4 / 96.2 & +2.4\% \\
Qwen3VL-4B & 4.44 / 8.27 & 52.4 / 99.9 & +1.2\% \\
Qwen3VL-8B & 8.77 / 16.33 & 58.3 / 111.2 & +0.7\% \\
InternVL3.5-1B & 1.06 / 1.98 & 46.5 / 88.7 & +4.4\% \\ \bottomrule
\end{tabular}
\end{table}

\subsection{Controlled analysis of the uncertainty components}
\label{subsec:intervention_appendix}

\textbf{Visual intervention.}
We first modify only the visual input $x$ while fixing context $t$. Specifically, for each image-question pair, we keep the question unchanged and progressively degrade the image using Gaussian blur with radii 0, 2, 4, 8, and 16. Then, we measure how $u_t$  and $u_{x \mid t}$ change. 
\cref{tab:intervention_a} shows that when we perturb the input image $x$, only $u_{x \mid t}$ increases monotonically, while $u_t$ remains fixed. This is because we might have lower confidence in a low-quality image, which broadens the possible answer set of the given $x, t$ pair.

\textbf{Context intervention.}
We modify the breadth of the answer space induced by $t$. Note that $u_{x \mid t}$ is also dependent on $t$; therefore, solely testing $u_t$ without changing $u_{x \mid t}$ is not possible. Instead, we construct a toy example where the direction of the uncertainty is predictable.

Let $s_i = p(m=1 \mid x, t, y_i)$ and $p(y \mid t) = \frac{1}{K}$. Then we have:
\begin{equation}
u_{x \mid t} = -\log (\text{CV}[S]^2 + 1) = -\log \frac{\mathbb E[S^2]}{\mathbb E[S]^2} = -\log\left[ K\frac{\sum_{i=1}^K s_i^2}{(\sum_{i=1}^K s_i)^2} \right].
\end{equation}

Now, let’s assume that exactly $r$ of the $K$ context-plausible candidates are plausible with the given input (\ie, $s_i = 1$ or $0$ and only $r$ scores are 1). For example, assume we have a question ``what color is this car among blue, dark blue, and red?'' and a dark blue-ish car image. In this case, $K=3$ (blue, dark blue, red) and $r=2$ (blue, dark blue). Under this simplified assumption, $\sum_{i=1}^K s_i^2 = r$ and $(\sum_{i=1}^K s_i)^2 = r^2$, thus, $u_{x \mid t} = \log r -\log K$.
By assuming $r=1$, \ie, only one plausible answer. We have $u_{x \mid t} = -\log K$. Namely, in this setting, $u_{x \mid t}$ becomes smaller (or more certain) by enlarging $K$. 
Note that this is not a realistic setting, but a controlled experiment to verify the behavior of each module. We will explore a more realistic setting later.

To approximate this controlled case, we select ``how many'' questions from VQA v2 whose majority answer is supported by at least nine human annotations. We change the question to ``\texttt{\{original question\}. Answer with a single integer from 0 to \{K-1\}}'', and vary $K$.

\cref{tab:intervention_b} shows the results. As we expected, $u_t$ is monotonically increasing because the plausible answer space becomes larger, and $u_{x \mid t}$ is monotonically decreasing as we derived previously. The results are computed using 525 samples.

Again, this is an unrealistic setting and does not represent natural open-ended settings. As the candidate set expands, it is likely to include additional plausible answers compatible with the input. For example, let's assume a ``what color'' question. Increasing the size of the answer space might lead to confusing answers, such as blue, dark blue, navy blue, and so on. In other words, in practice, $r$ is a function of $K$, i.e., $r(K)$, and increases with $K$, and $\log \frac{r(K)}{K}$ can be increased.

To examine a less restrictive and more natural controlled setting, we select ``what color'' and ``how many'' questions from VQA v2 whose normalized majority answer is supported by at least three human annotations. We select 1,809 samples from the 10k VQA v2 evaluation set. For each question, we construct nested answer sets of sizes $K=2, 4, 8$. Color distractors are sampled from a fixed color vocabulary, while count distractors are sampled from neighboring non-negative integers. In this more natural construction, expanding $K$ may also alter the distribution of matching probabilities.

For example, given the question ``How many men are wearing hats?'' with the ground-truth answer ``1'' and $K=8$, we construct:
``\texttt{How many men are wearing hats? Choices: A. 3, B. 4, C. 1, D. 6, E. 7, F. 2, G. 0, H. 5. Answer with the option letter only.}''
When $K=2, 4$, we simply use a subset of the above example with fewer choices and shuffled order.

We also evaluate the original open-ended question as a reference (``$-$'' in the table). Note that because the original open-ended questions do not have an explicitly specified $K$, this result is not directly comparable with the controlled K experiment, which is similar to MCQ settings.

\cref{tab:intervention_c} shows the average $u_t$, $u_{x\mid t}$, and $u$ in the selected samples. There are three interesting findings in this table.
First, on average, altering $K$ from 2 to 8 to have a broader answer space significantly increases the context-related uncertainty from $-0.546$ to $0.018$. We also observe that 96.6\% of samples show larger $u_t$ when $K=8$ compared to $K=2$.
Second, the original open-ended $t$ has $u_t$ of $1.786$, which is much larger than restricted contexts. This is expected because open-ended questions might have a larger number of plausible answers compared to questions limiting the number of answers.
Finally, $u_{x \mid t}$ increases as $K$ increases. In natural open-ended settings, expanding $p(y\mid t)$ often introduces semantically related variants that remain compatible with $x$, rather than only incompatible candidates. Therefore, we generally expect $u_{x\mid t}$ to increase because the relative mass of compatible answers can increase with $K$. This differs from the simplified $r=1$ case, where every additional candidate is assumed to be incompatible.

\subsection{Runtime analysis}
\label{sec:appendix_runtime}

We first show the additional parameter size and memory consumption by \ours in \cref{tab:additional_params_memory}.
In terms of the parameter size, \ours adds 46.5M to 58.3M parameters depending on the backbone, corresponding to a relative overhead of 0.7\% to 4.4\% depending on the backbone. This shows that our uncertainty module is lightweight.

\begin{wraptable}{r}{0.34\linewidth}
\centering
\caption{\small {\bf Runtime Analysis.} We report the average throughput of uncertainty estimation methods on the sampled VQA v2 dataset using the Qwen3VL-2B-Instruct model with \ours. The number of maximum generation tokens is set to 4096 in this experiment. Runtime is measured on a single RTX 3090 GPU.}
\label{tab:appendix_runtime}
\small
\begin{tabular}{ll}
\toprule
Methods & Throughput (s) \\ \midrule
Consistency & 34.87 \\
NLL & 5.95 \\
Verbalized Conf & 6.10 \\
P(Correct) & 6.01 \\
VL Uncertainty & 16.33 \\ 
\rowcolor{gray!20}
\ours (Ours) & \textbf{5.88} \\
\bottomrule
\end{tabular}
\vspace{-2em}
\end{wraptable}
We measure the actual runtime by various uncertainty estimation methods.
Note that the multimodal understanding benchmarks used in our experiments are usually sufficient with very short generated tokens (\eg, ``A'' for MCQ VQA tasks or ``A cat'' for open-ended VQA tasks). Therefore, we set the maximum generated tokens to 4,096 to examine the scenario when we need to generate a longer-context output; this could include the reasoning trace \cite{wei2022chain}, or long context generation.

\cref{tab:appendix_runtime} shows the results. We randomly select 100 samples from the VQA v2 dataset and measure the runtime for uncertainty estimation methods. In this experiment, \ours shows the most efficient inference time uncertainty estimation. Specifically, compared to sampling-based methods, such as consistency (we generate five diverse samples) and VL Uncertainty (we randomly generate five noises), \ours performs much faster inference, but achieves significantly better uncertainty estimation performance as shown in the main results.

Training takes about 3 L40 GPU hours after cache extraction.
Cache extraction took 5.5 hours on 40 L40 GPUs (220 GPU-hours) for 200k samples.
For evaluation, consistency or VL uncertainty takes more than a day with 1 L40 GPU when we have about 10k samples (\eg, VQA v2 or MMMU test); meanwhile, because \ours can re-use the cached features, once features are cached, it takes less than a minute to evaluate; cache extraction takes about 3-4 hours in the same setting.

\subsection{Visualization}
\label{sec:appendix_visualization}

\begin{figure}[t!]
\centering
\begin{subfigure}[b]{0.29\linewidth}
\includegraphics[width=\linewidth]{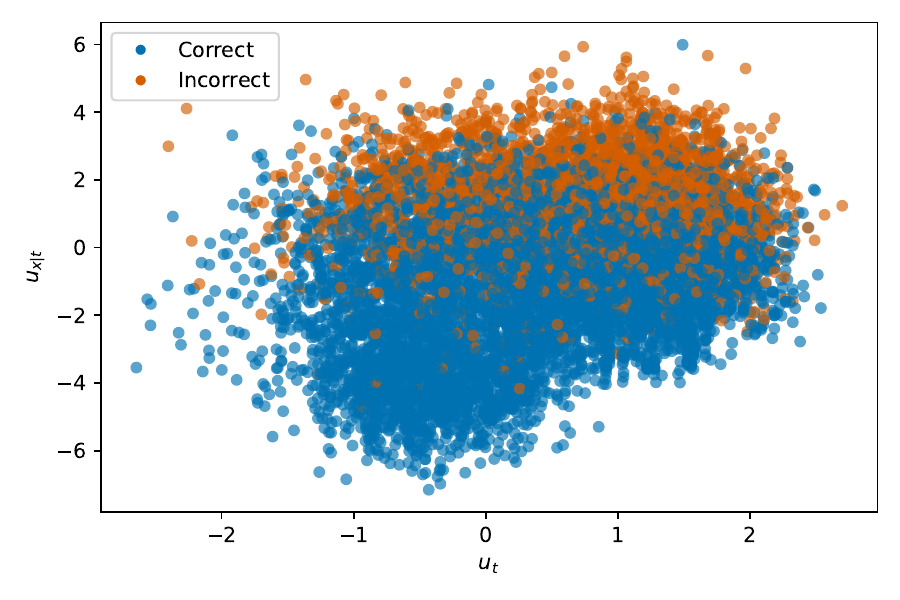}
    \caption{\small {\bf $u_t$ vs. $u_{x \mid t}$.}}
    \label{fig:appendix_u_t_vs_u_xt}
\end{subfigure}
\begin{subfigure}[b]{0.7\linewidth}
\includegraphics[width=\linewidth]{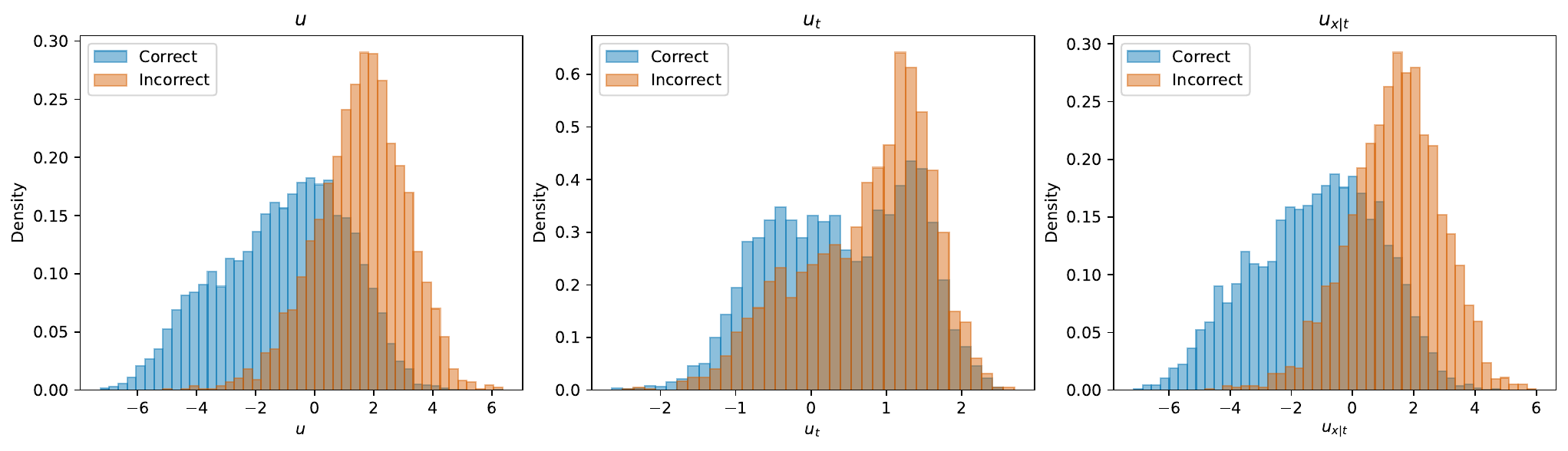}
\caption{\small {\bf Histogram of each uncertainty}}
\label{fig:appendix_u_histogram}
\end{subfigure}
\caption{\small {\bf Statistics of uncertainty values.} We plot the statistics related to $u$, $u_t$, and $u_{x \mid t}$ on the VQA v2 dataset using the Qwen3VL-2B-Instruct model. Blue samples denote correct samples, while orange samples denote incorrect samples.}
\label{fig:appendix_u_stats}
\vspace{-.5em}
\end{figure}

\cref{fig:appendix_u_stats} shows the statistics of uncertainties on VQA v2 datasets using the Qwen3VL-2B-Instruct model. Blue and orange samples correspond to correct and incorrect samples, respectively.

For the qualitative visualizations, we use \ours with the Qwen3VL-8B-Instruct model.
We visualize the samples having high and low uncertainty in each uncertainty term: \cref{fig:appendix_unc_vis1} shows the samples with our proposed uncertainty, \cref{fig:appendix_unc_vis2} shows the samples with context-specific uncertainty $u_t$, and \cref{fig:appendix_unc_vis3} shows the samples with multiplicity-specific uncertainty $u_{x \mid t}$.

In addition, we illustrate the samples having the same $x$ but different $t$ in \cref{fig:appendix_unc_vis_same_x}. As shown in the figure, the level of uncertainty can be changed a lot by choosing different context. For example, ``Is this food ready to eat'' can be answered easily, but ``Is this pizza homemade?'' is somewhat hard to answer; we need more information to answer the question.

We also visualize the samples having the same context (question) $t$ but different input (image $x$) in \cref{fig:appendix_unc_vis_same_t}. Even with the same $t$ (therefore, having the same $u_t$), the degree of uncertainty could vary a lot as shown in the figure.

Finally, we report the samples with opposing $u_t$ and $u_{x \mid t}$, \eg, $u_t$ is high but $u_{x \mid t}$ is low, and vice versa. \cref{fig:appendix_unc_vis_contrast} shows the samples. For example, even though a question can be answered in just two answers (``yes'' or ``no''; so $u_t$ is very low), it is somewhat difficult to answer the question ``Are all people about the same age?'' in the first example, resulting in a high uncertainty value due to $u_{x \mid t}$.

\begin{figure}[t!]
    \centering
    \includegraphics[width=\linewidth]{figures/unc_vis1.pdf}
    \caption{\small {\bf Visualization of the certain/uncertain samples by the proposed uncertainty scores $u$.}}
    \label{fig:appendix_unc_vis1}
\end{figure}

\begin{figure}[t!]
    \centering
    \includegraphics[width=.98\linewidth]{figures/unc_vis2.pdf}
    \caption{\small {\bf Visualization of the certain/uncertain samples by the context-specific uncertainty scores $u_t$.}}
    \label{fig:appendix_unc_vis2}
\end{figure}

\begin{figure}[t!]
    \centering
    \includegraphics[width=\linewidth]{figures/unc_vis3.pdf}
    \caption{\small {\bf Visualization of the certain/uncertain samples by the multiplicity-specific uncertainty scores $u_{x \mid t}$.}}
    \label{fig:appendix_unc_vis3}
\end{figure}

\begin{figure}[t!]
    \centering
    \includegraphics[width=\linewidth]{figures/same_image_different_t.pdf}
    \caption{\small {\bf Uncertainty can vary by context $t$ even with the same input $x$.}}
    \label{fig:appendix_unc_vis_same_x}
\end{figure}

\begin{figure}[t!]
    \centering
    \includegraphics[width=.95\linewidth]{figures/same_t_different_x.pdf}
    \caption{\small {\bf Uncertainty can vary by the relationship between $x \mid t$ and the set of plausible answers $y$ (\ie, their multiplicity), even with the same input $t$.}}
    \label{fig:appendix_unc_vis_same_t}
\end{figure}

\begin{figure}[t!]
    \centering
    \includegraphics[width=\linewidth]{figures/contrast_vis.pdf}
    \caption{\small {\bf Samples with opposing uncertainty components.}
    Top: low $u_t$ and high $u_{x|t}$. Bottom: high $u_t$ and low $u_{x|t}$. These examples illustrate cases where task-only and image-conditioned uncertainty provide conflicting signals.}
    \label{fig:appendix_unc_vis_contrast}
\end{figure}

\section{Discussion and Limitation}
\label{sec:limitation}

\paragraph{Limitation in theory.}
In theory, we use the uniform assumption on $p(y \mid t)$. Although we believe that it is reasonable and not strong assumptions, this does not generalize to case where $p(y \mid t)$ is not a uniform distribution. However, our assumption can be viewed as a relaxed version of those cases; furthermore, the empirical evidence supports that our formulation is suitable in many scenarios, including multimodal understanding, error detection, and multimodal MCQ tasks.

\paragraph{Limitation in method design.}
Although our formulation is model-agnostic and can be generalized to any form of multimodal tasks, our actual method design has some limitations. 

The first limitation of \ours in terms of method design is the approximation of $p( y \mid t)$, or specifically $K_t$. However, we believe that the approximation is one of the most reasonable and sound ways considering that the true $p(y \mid t)$ is impossible to access. We also use a control parameter $\lambda$ to control the approximation error induced by this approximation; our experiments show that $\lambda=1$ (\ie, using the approximation without any control) actually performs well on average (\cref{tab:lambda_ablation}); we chose $\lambda=0.25$ because it performs the best in the hyperparameter selection criteria (\ie, MMMU validation), but in practice, we believe that $\lambda=1$ is also acceptable, considering the empirical evidence in our parameter study.

Another limitation is the quality of the matching probability module, \ie, MLLM-as-verifier. As shown in \cref{fig:appendix_matching_probabilty}, the MLLM-as-verifier is not a perfect estimator; the quality of matching probability estimator can degrade the performance of \ours as shown in \cref{sec:appendix_why_no_train_matching_probability}. This makes the framework simple and broadly applicable, but the resulting uncertainty targets can inherit calibration errors or semantic biases from the verifier. We partially address this issue through validation-based calibration, but a more principled calibration of semantic matching probabilities remains an interesting direction for future work.

\paragraph{Potential extensions.} 

Although this paper focuses on open-ended multimodal VQA, the proposed formulation is not limited to this setting. More broadly, $X$ can denote an observed input, $T$ can denote the task context or environment specification that constrains the valid outputs, and $Y$ can denote a candidate response or action.
For example, in agentic AI settings, uncertainty may arise from ambiguity in the user request, the environment state, or the set of valid actions \cite{oh2026uncertainty}. In this case, $X$ can denote the observed state or user input, $T$ can denote the task context or environment specification, and $Y$ can denote a candidate response or action.
Similarly, for reasoning models, the context $t$ can include an intermediate reasoning trace that define the space of acceptable outputs. We leave a systematic study of these broader settings to future work.

\section{Broader impacts}
\label{sec:broader_impacts}
This work aims to provide a better multimodal predictive uncertainty of MLLMs, which potentially supports safer deployment in high-stakes scenarios. We do not anticipate direct negative impacts specific to \ours beyond those of uncertainty-aware MLLMs in general.

We use VizWiz \cite{vizwiz} to evaluate both open-ended VQA and unanswerability detection. Its images and spoken questions were contributed by blind users seeking help with everyday visual tasks; the reference answers were collected from crowd workers. A question may be unanswerable from the submitted image even though the user still needs an answer. If an assistive system gives an incorrect answer with low uncertainty, the user may act on information they cannot easily verify. If it abstains when the image does contain enough information, the user may have to retake the image or ask someone else for help. Evaluating such a system with blind and low-vision users would help determine when abstention is useful and what assistance to offer next.

These errors also have different consequences in other high-stakes settings. For example, a falsely confident answer about a medication label could lead someone to take the wrong dose, while unnecessary abstention could delay access to instructions that were legible. Our benchmarks measure how well uncertainty scores distinguish correct from incorrect predictions; they do not establish when it is safe to act on an answer or when a system should abstain. Before using \ours for such decisions, both types of error should be evaluated under the intended conditions of use, with an alternative way to obtain or verify an answer when needed.

\makeatletter
\makeatother
\end{document}